\documentclass[10pt,twocolumn,letterpaper]{article}

\usepackage[pagenumbers]{iccv} 

\newcommand{\red}[1]{{\color{red}#1}}

\definecolor{iccvblue}{rgb}{0.21,0.49,0.74}
\usepackage[pagebackref,breaklinks,colorlinks,allcolors=iccvblue]{hyperref}

\def\paperID{2101} 
\def\confName{ICCV}
\def\confYear{2025}

\usepackage{booktabs}
\usepackage{multirow}
\usepackage{color, colortbl}
\usepackage{subcaption}
\usepackage{tabu}
\usepackage{pifont}
\usepackage{makecell}
\usepackage{paralist}
\usepackage{threeparttable}
\usepackage{enumitem}
\usepackage{hhline}
\usepackage{scalerel,xparse}

\definecolor{darkblue}{RGB}{94,110,186}
\definecolor{Gray}{gray}{0.5}

\newcommand{\gray}[1]{\textcolor{gray}{#1}}
\newcommand{\darkblue}[1]{\textcolor{darkblue}{#1}}
\definecolor{darkGreen}{RGB}{92, 148, 110}
\definecolor{asparagus}{RGB}{0.53, 0.66, 0.42}
\newcommand{\darkGreen}[1]{\textcolor{darkGreen}{#1}}
\newcommand{\cmark}{\ding{51}}%
\newcommand{\xmark}{\ding{55}}%

\newlength\savewidth\newcommand\shline{\noalign{\global\savewidth\arrayrulewidth
  \global\arrayrulewidth 1pt}\hline\noalign{\global\arrayrulewidth\savewidth}}

\makeatletter\renewcommand\paragraph{\@startsection{paragraph}{4}{\z@}
  {.5em \@plus1ex \@minus.2ex}{-.5em}{\normalfont\normalsize\bfseries}}\makeatother

\title{Make Your Training Flexible: Towards Deployment-Efficient Video Models}

\def\ModelName{FluxViT}
\def\methodname{Flux}

\newcommand\blfootnote[1]{%
  \begingroup
  \renewcommand\thefootnote{}\footnote{#1}%
  \addtocounter{footnote}{-1}%
  \endgroup
}

\author{
    Chenting Wang$^{1,2}$\quad
    Kunchang Li$^{2,3}$\quad \\
    Tianxiang Jiang$^{2,4}$\quad
    Xiangyu Zeng$^{2,5}$\quad
    Yi Wang$^{2\heartsuit}$\quad
    Limin Wang$^{2,5\heartsuit}$\vspace{0.2em}\\
    \footnotesize{$^1$Shanghai Jiao Tong University}
    \quad $^2$Shanghai AI Laboratory \quad $^3$University of Chinese Academy of Sciences\\
    \footnotesize{$^4$University of Science and Technology of China\quad
    $^5$State Key Laboratory for Novel Software Technology, Nanjing University}
}

\begin{document}
\maketitle
\begin{abstract}
Popular video training methods mainly operate on a fixed number of tokens sampled from a predetermined spatiotemporal grid, resulting in sub-optimal accuracy-computation trade-offs due to inherent video redundancy. They also lack adaptability to varying computational budgets for downstream tasks, hindering applications of the most competitive model in real-world scenes. We thus propose a new test setting, Token Optimization, for maximized input information across budgets, which optimizes the size-limited set of input tokens through token selection from more suitably sampled videos. To this end, we propose a novel augmentation tool termed Flux. By making the sampling grid flexible and leveraging token selection, it is easily adopted in most popular video training frameworks, boosting model robustness with nearly no additional cost. We integrate Flux in large-scale video pre-training, and the resulting \textbf{FluxViT} establishes new state-of-the-art results across extensive tasks at standard costs. Notably, with 1/4 tokens only, it can still match the performance of previous state-of-the-art models with Token Optimization, yielding nearly 90\% savings. All models and data are available at \url{https://github.com/OpenGVLab/FluxViT}.
\end{abstract}    
\section{Introduction}
\label{sec:intro}

\begin{figure}[t]
    \centering
    \includegraphics[width=0.85\linewidth, keepaspectratio]{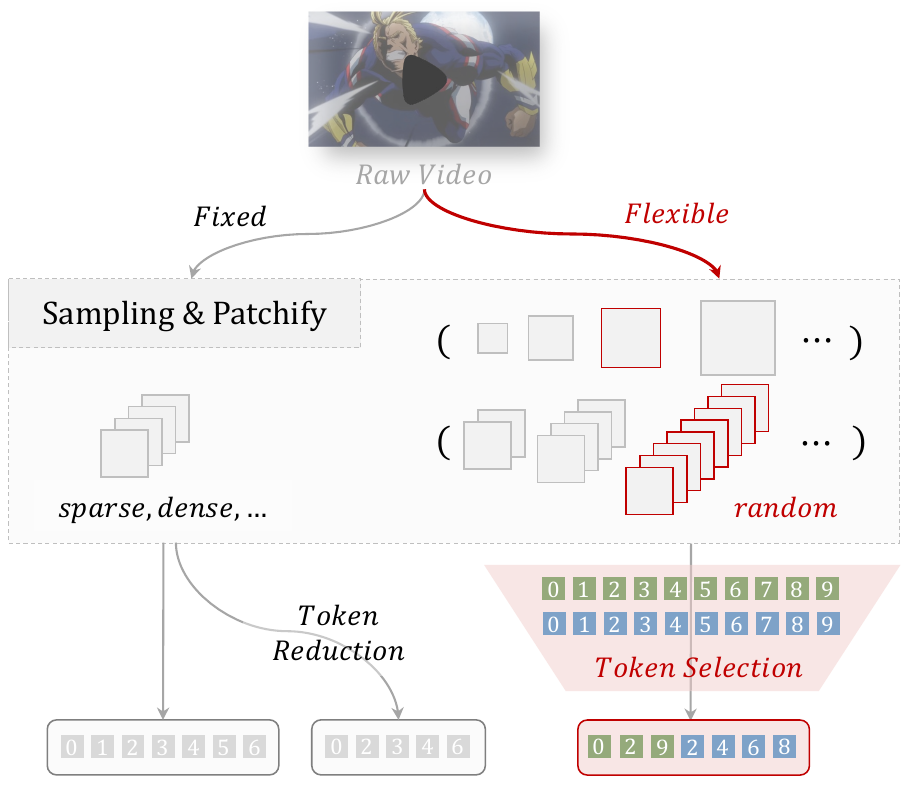}
    \vspace{-0.4cm}
    \caption{
    \textbf{Flux} (right) employs flexible sampling and token selection to achieve Token Optimization. Common methods(left) use rigid sampling(and use token reduction for applications directly).
    }
    \vspace{-0.4cm}
    \label{fig:hero}
\end{figure}

\blfootnote{$\heartsuit$ Corresponding authors.}

Video representation learning~\cite{tsn, tsm, slowfast} is a fundamental task in computer vision, as it is critical in developing spatiotemporal perception and procedural reasoning in multimodal large language models~\cite{videochat, videochatgpt, blip2, iv2} and embodied AI systems~\cite{vln, scalevln}. However, the deployment of video models is computationally intensive due to the high volume of video tokens compared to image tokens. Popular video training methods work on fixed-count patched tokens sampled from a predefined spatiotemporal grid (e.g., 8$\times$224$^2$), leading to considerable redundancy in training and deployment, especially for long, high-resolution videos. Moreover, this inflexible and fixed training achieves sub-optimal results for tasks that require dynamic resolution settings, including restricted computation or emphasizing frame counts rather than frame resolution. Researchers have developed different approaches for better trade-offs.

First, to fit a limited computation budget in inference, many works~\cite{vidtldr, Qing2022mar, Hwang2022EVERESTEM} use token reduction on the same densely sampled tokens as in training, showing sub-optimal performance. For one thing, there is a trade-off between the complexity (cost) and performance of the token reduction strategy. Current strategies can not perform well with a significant reduction rate. For another, the inflexibly trained models can not generalize with the sparse, masked tokens.

Second, existing methods for flexible network training, such as Resformer~\cite{resformer} and FFN~\cite{frameflexiblenet}, have demonstrated the efficacy of flexible networks operating at different spatial or temporal resolutions but not both simultaneously. Although these approaches achieve robust performance across diverse settings, we contend that merely reducing frame counts or resolution under computational constraints fails to optimize token capacity utilization and remains susceptible to redundancy. Notably, FFN exhibits substantial performance degradation when operating with only two frames in their research. We argue that the input token set with a size equal to 2 frames can be optimized. Also, they are not validated in large-scale pre-training to achieve competitive results for real-world applications, and the performance gain with larger resolution input comes with quadratic computation cost increase, which is also insufficient.

We propose a new perspective of the desired optimal computation-and-accuracy and spatial-and-temporal trade-offs across various settings, termed \textbf{`Token Optimization'} (TO in short), as shown in Figure \ref{fig:hero}. It involves selecting an optimized set of input tokens based on budget from more properly sampled videos for information maximization. Since there is no perfect token selector for all settings, we promote a better solution of using flexible sampling, with more densely and finely sampled videos for higher computation and more sparsely sampled videos for lower budget. Spatial and temporal trade-offs are also achieved.

\begin{figure}[t]
    \centering
    \includegraphics[width=\linewidth, keepaspectratio]{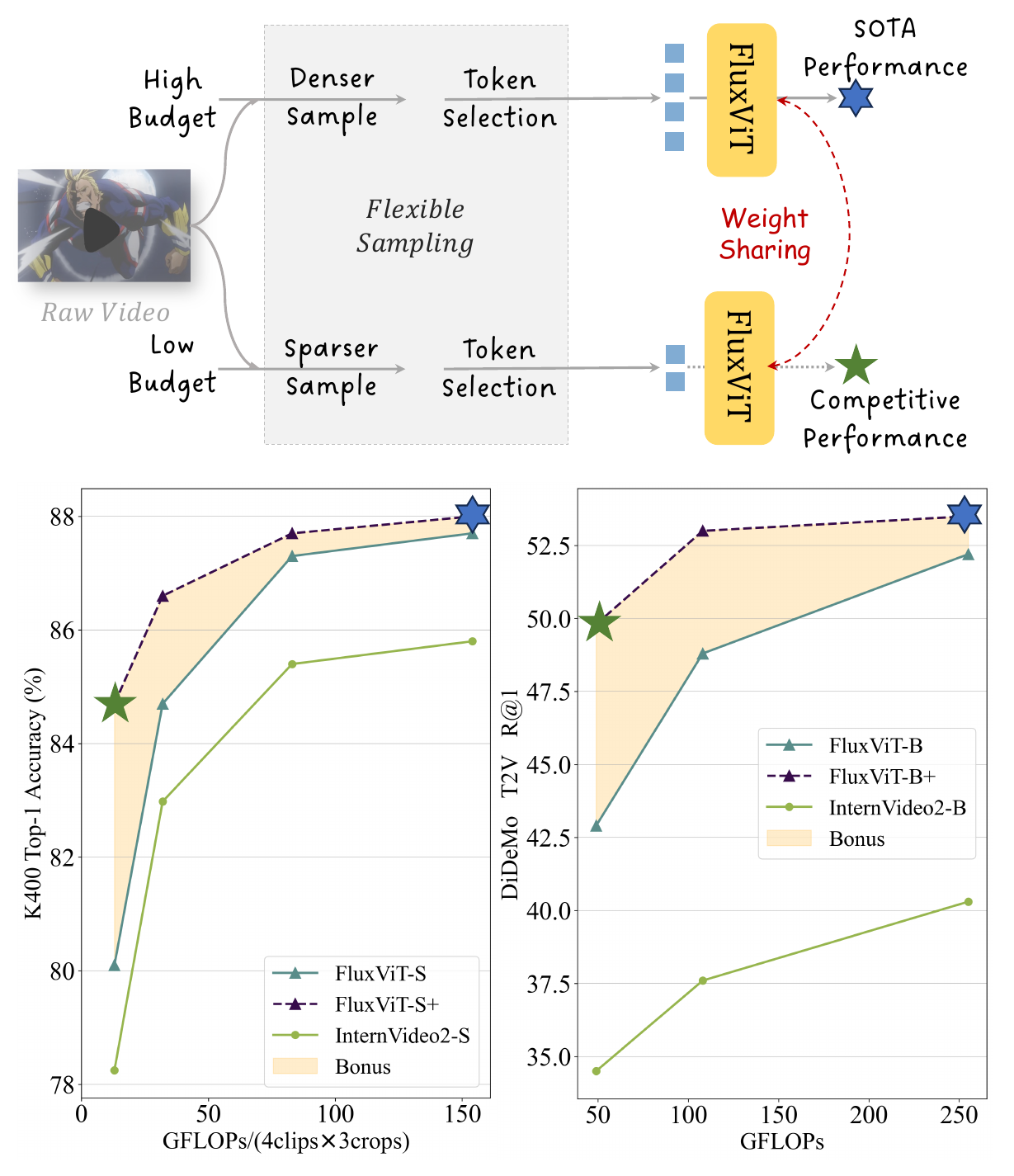}
    \vspace{-0.8cm}
    \caption{
    \textbf{Overview of our Flux method.}
    The same-scaled \ModelName{} and InternVideo2~\cite{iv2} series models are both pre-trained with the InternVideo2-1b model as the teacher using the same dataset. The ``FluxViT+'' refers to the results using Token Optimization at test time with the same GFLOPS.
    }
    \vspace{-0.4cm}
    \label{fig:meme}
\end{figure}

To better achieve TO, we propose a new data augmentation tool for training flexible video models, termed \textbf{Flux} for {\textbf{Fl}}exibility in {\textbf{u}}niversal conte{\textbf{x}}ts, by utilizing the combination of flexible sampling and token selector to make training flexible while not introducing costs. We integrate it with the Unmasked Teacher~\cite{umt} pre-training, supervised-training, multi-modal contrastive training, and chat-centric training to validate its efficiency. Our new recipe brings more robustness caused by Flux augmentation both in standard settings and our TO setting. Two plug-in modules are used for better feature extraction regarding the reduced token set with flexible reduction rates and input spatiotemporal resolutions. We build bottom-up ablation studies in turning the model flexible. The resulting model FluxViT achieves supreme improvements regarding the currently state-of-the-art InternVideo2-Dist series models in fair settings across budgets and across tasks, as in Figure \ref{fig:meme}.

Our \ModelName{}-S outperforms the previous state-of-the-art small-scale model, InternVideo2-S~\cite{iv2}, by \textbf{2.2\%} on K400 under standard computation constraints. Moreover, it achieves comparable performance to InternVideo2-S while utilizing only about \textbf{10\%} of inference cost. \ModelName{}-B achieves competitive results with state-of-the-art much larger scale models on various tasks, including scene-based action recognition task~\cite{k400} (\textbf{90.0\%} top-1 accuracy on K400), motion-intensive~\cite{goyal2017something} (\textbf{75.8\%} top-1 accuracy on SSv2), long-term task~\cite{coin} (\textbf{94.1\%} top-1 accuracy on COIN), zero-shot text-to-video retrieval~\cite{msrvtt, didemo, activitynet, lsmdc, msvd} (\textbf{49.9\%} R@1 on MSRVTT, \textbf{53.5\%} R@1 on DiDeMo and \textbf{56.7\%} R@1 on ActivityNet). Notably, it maintains state-of-the-art performance among the similar scaled models while using only \textbf{1/4} to \textbf{1/2} of their tokens, resulting in \textbf{70\%} to \textbf{95\%} savings in computation. On chat centric tasks including MVbench~\cite{videochat} for general perception and Dream-1k~\cite{tarsier} for fine-grained caption, our model surpasses widely used SigLIP, CLIP and UMT in `linear prob' setting across computation budgets, where only the projector between the ViT and LLM is trained for comparability regarding the ViT's capability as the common stage-1 training in chat models. This highlights Flux's strong effect across computation constraints and tasks and underscores the robustness afforded by such flexibility.
\section{Related Work}

\paragraph{Flexible Model Training.}

Many recent studies focus on developing a method to train a single model that performs well across various settings, facilitating seamless adaptation to diverse downstream deployment scenarios. These methods include training a model as a set of shared weights to be pruned for specific tasks~\cite{yu2020bignas, onceforall}, as well as optimizing models for varied spatial resolutions~\cite{resformer}, frame counts \cite{frameflexiblenet}, spatiotemporal resolutions~\cite{liu2024oryx, wang2024qwen2, liu2023improvedllava, multigrid}, patch sizes \cite{flexivit}, and representation dimensions~\cite{Kusupati2022MatryoshkaRL, cai2024matryoshkamllm}. The Resformer~\cite{resformer} and FFN~\cite{frameflexiblenet} methods propose co-training with three different resolution settings in a single batch, utilizing self-distillation mechanisms to enhance performance under limited input resolutions for spatial and temporal dimensions, although separately. We argue that they can't achieve optimal performance within constrained computational budgets. Simply reducing frame counts and resolutions is suboptimal. Further, performance gains brought by high-resolution inputs lead to significant cost increases, a challenge also noted in Oryx~\cite{liu2024oryx}. Token Optimization addresses this by optimizing the set of input tokens anytime.

\paragraph{Masked Vision Modeling.}

Mask has always been seen as a critical method in training and utilizing scaled vision models, especially in the video domain. Previous methods can be mainly divided into two categories. First, drawing insights from the masked language modeling~\cite{lu2019vilbert,dong2019unified,devlin2018bert}, researchers use masks in vision transformer~\cite{vit} pre-training as a data augmentation strategy, as seen in masked autoencoder frameworks~\cite{mae,videomae,videomaev2,st_mae} and BERT-like mask-then-predict frameworks~\cite{beit,beitv2,ibot,bevt}. Works like UMT~\cite{umt}, MVP~\cite{mvp}, and Milan~\cite{milan} further utilize such augmentation strategy to predict the CLIP~\cite{clip} model's feature instead of native pixels as a Teacher-Student pre-training method. Second, methods like MAR~\cite{Qing2022mar} and MCM~\cite{Hwang2022EVERESTEM} apply masking techniques during the fine-tuning stage and inference to save computation with minimal performance loss. We here propose that mask can be used as novel data-augmentation tool with our flexible sampling to boost performance with our Token Optimization idea.
\section{Method}
\label{sec:method}

\begin{figure}[t]
    \centering
    \includegraphics[width=\linewidth, keepaspectratio]{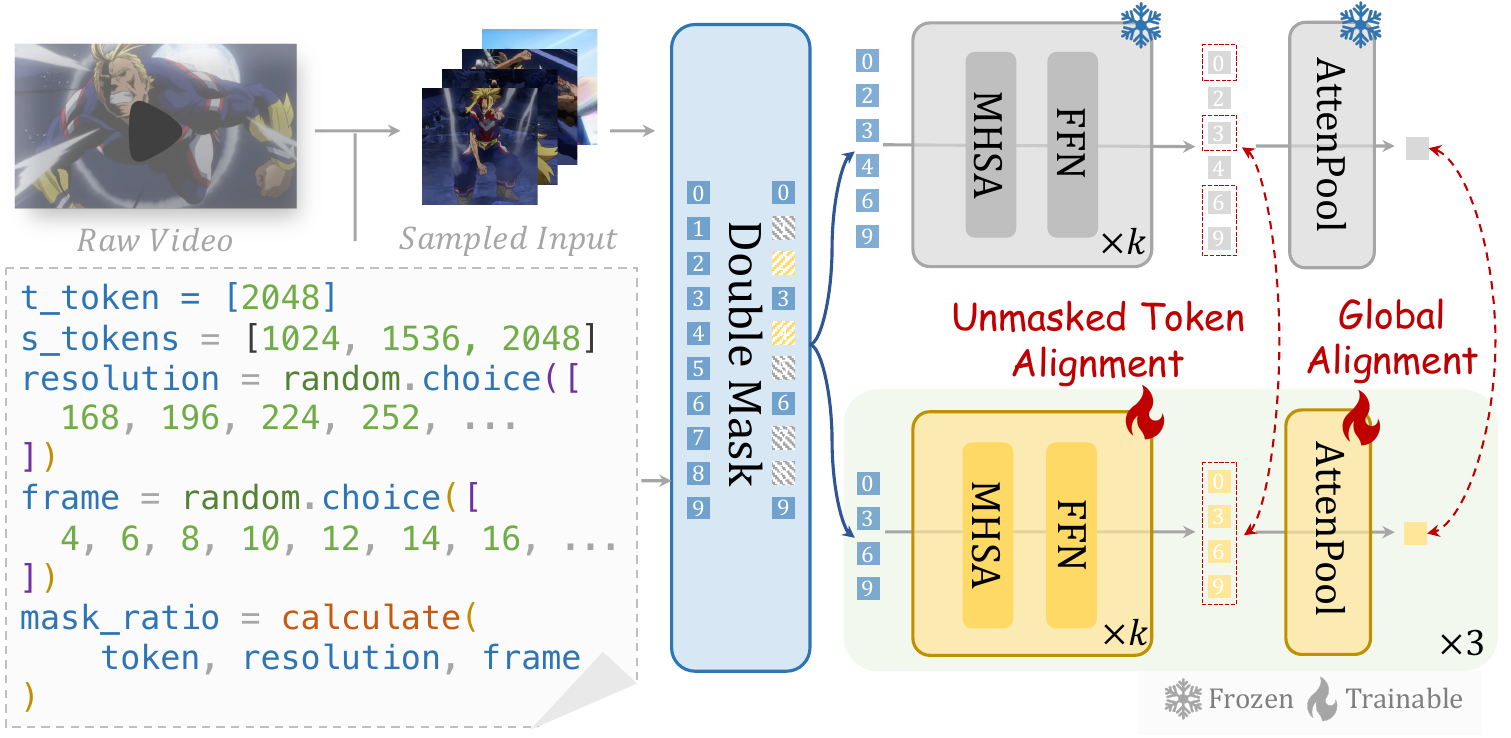}
    \vspace{-0.6cm}
    \caption{
    \textbf{Our proposed Flux method with UMT framework.}  We show that our proposed Flux training is easy to integrate with mainstream video training frameworks.
    }
    \vspace{-0.3cm}
    \label{fig:framework}
\end{figure}

In this section, we introduce our \textbf{\methodname{}} for effectively pre-training video models that represent videos at varying resolution scales using tokens with flexible quantities robustly.

\subsection{Flux-Pretraining}
\label{sec:fts}
We begin with a widely adopted pre-training paradigm in which the representations of a teacher model are employed as soft labels for a student model. Popular approaches following this typically rely on fixed sampling grids and utilize all extracted visual tokens for both the teacher and student models. However, this methodology suffers from substantial redundancy, particularly in video data at finer resolutions. To address this, Unmasked Teacher (UMT) \cite{umt} proposes using only a subset of the visual tokens for the student while retaining the full input for the teacher. This masked alignment strategy enhances both efficiency and robustness for the student model. We propose a further examination of the input tokens for the teacher and leverage flexible sampling grids combined with a group-dynamic token selection mechanism. This enables us to select the same number of tokens as the original setting but from a larger pool extracted at flexibly higher spatiotemporal resolutions that are more informative, thus harvesting more diversified and meaningful representations from the teacher without cost, as shown in Figure \ref{fig:framework}. The other settings, like alignment losses, are the same as the original UMT, which trains the competitive InternVideo2-Series models~\cite{iv2}.

\begin{figure}[tp]
    \centering
    \includegraphics[width=\linewidth, keepaspectratio]{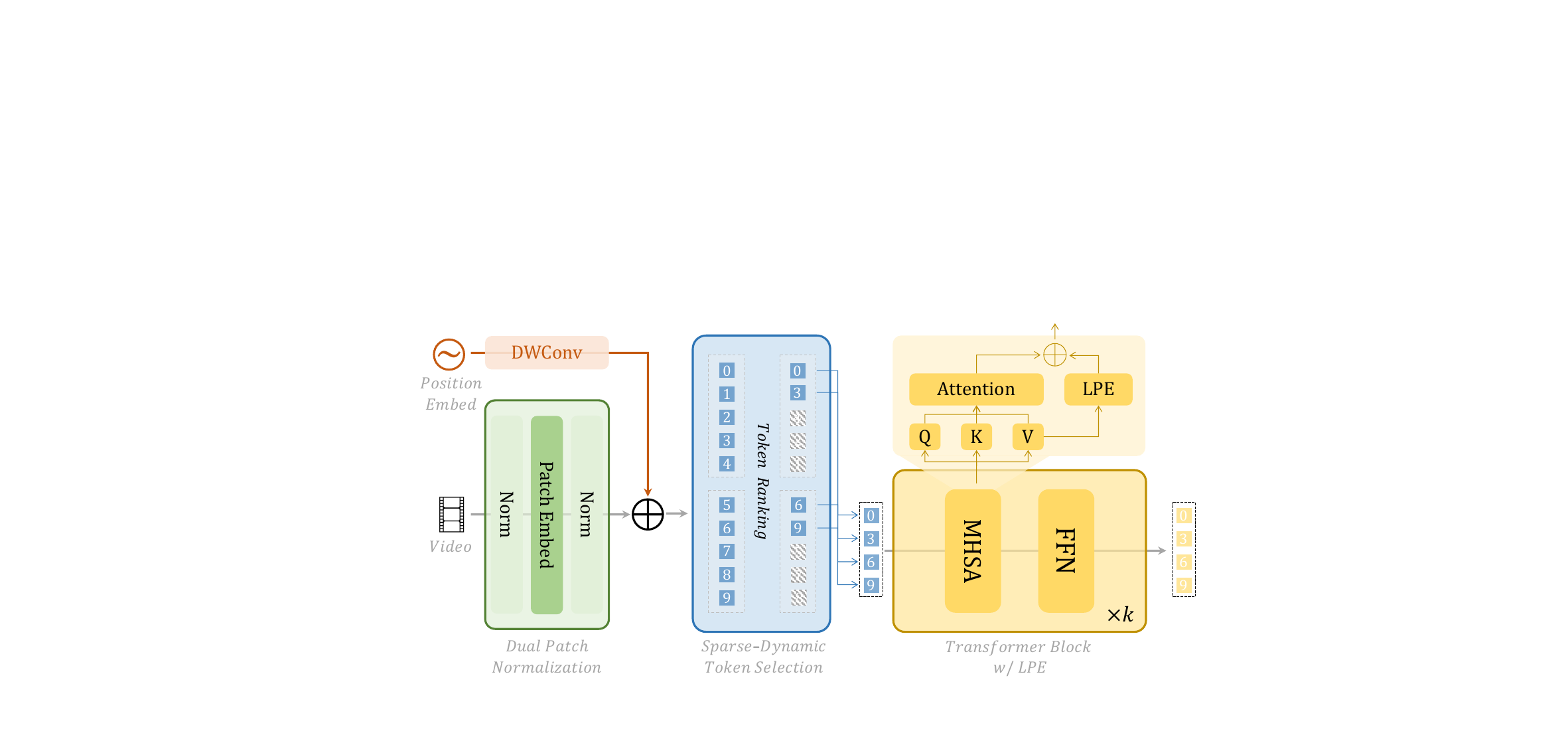}
    \vspace{-0.6cm}
    \caption{
    \textbf{Our proposed essential modules for Flux.} From the model side, Flux modules include Group-dynamic token selector, dual patch norm, and Global-Local positional embedding.  
    }
    \vspace{-0.4cm}
    \label{fig:framework_demo}
\end{figure}

\paragraph{Flexi-Sampling.}
For each video in a batch, we randomly select a frame count ranging from \( F_{min} \) to \( F_{max} \) with a temporal stride \( t_s \) and select a spatial resolution between \( R_{min} \) and \( R_{max} \) with a spatial stride \( r_s \) and enforce a threshold \( T_{thres} \) in keeping a reasonable size of visual token pool. 

\paragraph{Double Mask Module.}
To make the training flexible while not introducing new costs, we leverage token selection in the teacher input side. We propose a simple but effective method that prioritizes tokens exhibiting significant changes between frames. We use additional sparse groups for this selection to ensure coverage of the full video and resist several rapid changes in videos, termed \textbf{Group-Dynamic Selector}. Formally, with a pre-defined token count $K$, which is set to keep unchanged teacher forward cost, and a given tokenized frame sequence \{$F_1$, $F_2$, ..., $F_T$\} by teacher's patch embedding layer, we uniformly sample it into $N$ sparse groups where $B_i=\{F_{t} \mid t \in [iB, (i+1)B-1)\}, i \in [0, N)$. In each group, we compute the dynamic value as $\mathrm{D}(F_{t+1, i})=||F_{t+1, i} - F_{t, i}||_p$, where $F_{t, i}$ means the value of $i_{th}$ spatial token in the $t_{th}$ frame and $p$ means the $p$-dim distance. Then we take the $\mathrm{top}(\frac{K}{N})$ tokens with the highest dynamic values in each group to form $K$ visible tokens in total. This selector, along with flexible sampling, is utilized for the teacher model. We also employ another student mask based on the teacher's attention score of CLS token adopted in Unmasked Teacher. The usage of the two masks is referred to as the Double Mask Module. Such a module greatly boosts the student model's performance and brings no additional cost, showing a natural integration of Flux in the UMT framework. 

\subsection{Flux-Tuning}
For the stage of supervised tuning, we keep the flexi-sampling method and group-dynamic selector for the trained student model. Table \ref{tab:ablations_multi_k400_iv2} shows such a method's gain even on a competitive model pre-trained with standard setting, showing Flux as a general augmentation tool.

\paragraph{Token Optimization}
As proposed in the introduction, we hope to maximize token information under any given budget, termed as Token Optimization at downstream tasks. Thus, during evaluation, we use a subset of the train-set to find optimized sampling under a given token count for the task. We provide a heuristic searching method in the Appendix, which takes acceptable overhead.

\subsection{FluxViT}
\label{sec:fluxvit}
Conventional ViT architecture struggles with the variable token numbers, flexible spatiotemporal resolution, and sparsity patterns inherent in Flux training. Thus, we propose two useful plug-in augmentation modules:
Global-Local Positional Embedding (\textbf{GLPE}) and Dual Patch Normalization (\textbf{DPN}), as shown in Figure~\ref{fig:framework_demo}.

\paragraph{Global-Local Positional Embedding.} 
As we utilize flexible sampling with various spatiotemporal resolutions and token reduction rates, it is important to encode the position information of the selected input tokens about where they are from and how they are related. Thus, Global-Local Positional Embedding (GLPE) is utilized. We enhance the global learnable positional embedding, initialized using the sine-cosine method for the maximum possible input size, by applying a Depth-Wise Convolution~\cite{mobilenetv1}. Then, to encode local positional information at a finer granularity in the position-invariant attention mechanism, we apply a Linear Projection function (\textbf{LPE}) to the value vector $\mathbf{V}$. The modified attention is formulated as:
\begin{equation}
\mathbf{Z} = (\mathrm{Softmax}(\frac{\mathbf{QK}^T}{\sqrt{D}})+LPE )\mathbf{V}.
\end{equation}
Such LPE can be generated regardless of the input token number by adding a bias term for the permutation-invariant attention. This is vital in Flux to encode the discrete position information. The Value-dependent nature of LPE is more robust in Flux than others, like RoPE, in that the unmasked tokens are from different places from the source. The encoding of the relative relations between them is naturally Value-dependent. RoPE is more useful in scenarios without token selection for length extension. 

\paragraph{Dual Patch Normalization.}
The Patch Embedding layer plays a crucial role in dynamics estimation and it must be robust to various spatiotemporal resolution distributions. We observed that an additional Layer Normalization after the standard Patch Embedding improves the accuracy of dynamic estimation, leading to more effective token selection. Furthermore, the flexible sampling strategy leads to a varied distribution of input tokens. Consequently, we find the gradient norm of the Patch Embedding layer can become excessively large, as shown in the Appendix. To stabilize training, we introduce a second Layer-Normalization layer before the Patch Embedding operation. 

\subsection{Train with Varied Input Lengths}
\label{sec:variedls}
As we aim to train a flexible video model that can seamlessly adjust to varied token numbers according to budgets, we propose to use a multi-number co-training approach. For Flux-UMT frameworks, we suggest utilizing three distinct input numbers within a single batch for the student model. This strategy maximizes the benefits of the extensive teacher-forward computation while enhancing robustness under varying conditions. Specifically, we forward the student three times using three different token numbers and compute the teacher-student alignment loss for each. In standard training frameworks without a pre-trained teacher, we incorporate additional self-distillation, using the final aggregated features with the largest input as the teacher for the intermediate and the intermediate as the teacher for the smallest. We include training efficiency in the ablation.
\section{Experiments}
\label{sec:fluxpt}

We leverage InternVideo2-1B~\cite{iv2} as the teacher in Flux-UMT due to its strong performance in general video understanding tasks. We use the same data-decomposition and similar training recipe as the currently most competitive InternVideo2-series distilled models as baselines. Note that they are also built using the Unmasked Teacher framework under the supervision of InternVideo2-1B. For notation simplification, we use Flux-Single for Flux training with a single token count and Flux-Multi for Flux training with multi token counts, while Flux-PT is for Flux with UMT pretraining and Flux-FT for supervised tuning.

\subsection{Ablation Study}
\label{sec:ablation}
In the ablation study, we ablate the effect of our Flux through a systematic approach, progressively transforming the original UMT framework into our Flux-UMT and getting our competitive FluxViT. We utilize K710 for pre-training, which integrates the K400 \cite{k400}, K600 \cite{k600}, and K700 \cite{k700} datasets with duplicates removed. Performance is evaluated using scene-based K400 and motion-intensive SSv2~\cite{goyal2017something}. We use 1 clip $\times$ 1 crop for test configurations. For flexible sampling, we set the parameters as follows: $F_{min}$=4, $F_{max}$=24, $t_{s}$=2, $R_{min}$=168, $R_{max}$=252, $r_s$=28 and \( T_{thres} \)=(2048, 4096) by default.

\begin{table}[t]
    \centering
    \setlength\tabcolsep{2pt}
    \resizebox{0.9\linewidth}{!}{
    \begin{tabular}{l|r|c|c|c|c|c}
    \Xhline{1.0pt}	
    \multirow{2}{*}{\textbf{Method}} & \multirow{2}{*}{\textbf{Input Size}}&\multicolumn{4}{c|}{\textbf{Test Token Number}} & \multirow{2}{*}{\textbf{Avg}} \\ 
    ~ & ~ & 3072 & 2048 & 1024 & 512 & ~ \\
    \Xhline{1.0pt}	
    ~ & 2$\times$224$^2$ & $\varnothing$ & $\varnothing$ & $\varnothing$ & 69.5 & 69.5\\
    ~ & 4$\times$224$^2$ & $\varnothing$ & $\varnothing$ & 80.5 & 73.6 & 77.1 \\
    ~ & \textbf{8$\times$224$^2$} & $\varnothing$ & 83.9 & 81.9 & 72.7 & 79.5 \\
    \multirow{-1}{*}{8$\times$224$^2$} & 12$\times$224$^2$ & 84.6 & 84.2 & 80.9 & 69.8 & 79.9 \\
    \multirow{1}{*}{Direct Tuned} & 16$\times$224$^2$ & 84.5 & 83.5 & 79.1 & 67.0 & 78.5 \\
    ~ & 20$\times$224$^2$ & 84.1 & 82.7 & 77.7 & 64.3 & 77.2 \\
    ~ & 24$\times$224$^2$ & 83.6 & 82.1 & 76.4 & 62.3 & 76.1 \\ \cline{2-7}
    ~ & \textbf{Avg} & 84.2 & 83.3 & 79.4 & 68.5 & - \\ \cline{3-7}
    ~ & \textbf{Max} & 84.6 & 84.3 & 81.9 & 73.6 & - \\ 
    \Xhline{0.8pt}	
    ~ & 2$\times$224$^2$ & $\varnothing$ & $\varnothing$ & $\varnothing$ & 68.1 & 68.1\\
    ~ & 4$\times$224$^2$ & $\varnothing$ & $\varnothing$ & 79.9 & 74.7 & 77.3 \\
    ~ & 8$\times$224$^2$ & $\varnothing$ & 84.3 & 82.7 & 76.7 & 81.2 \\
    ~ & 12$\times$224$^2$ & 85.1 & 85.1 & 82.8 & 75.8 & 82.2 \\
    2048 fixed count & 16$\times$224$^2$ & 85.4 & 85.2 & 82.6 & 75.3 & 82.1 \\
    Flux-Single Tuned & 20$\times$224$^2$ & 85.6 & 85.1 & 82.2 & 74.6 & 81.9 \\
    ~ & 24$\times$224$^2$ & 85.4 & 84.8 & 82.0 & 74.1 & 81.6 \\ \cline{2-7}
    ~ & \textbf{Avg} & 85.4 & 84.9 & 82.0 & 74.2 & - \\ \cline{3-7}
    ~ & \multirow{2}{*}{\textbf{Max}} & 85.6 & 85.2 & 82.8 & 76.7 & - \\ 
    ~ & ~ & \darkGreen{$\uparrow$\textbf{1.0}}  & \darkGreen{$\uparrow$\textbf{0.9}}  & \darkGreen{$\uparrow$\textbf{0.9}} & \darkGreen{$\uparrow$\textbf{3.1}} & - \\ 
    \Xhline{1.0pt}	
    ~ & 2$\times$224$^2$ & $\varnothing$  & $\varnothing$ & $\varnothing$ &  69.3 & 69.3 \\ 
    ~ & 4$\times$224$^2$ & $\varnothing$ & $\varnothing$ & 80.5 & 72.8 & 76.7 \\ 
    ~ & 8$\times$224$^2$ & $\varnothing$ & 83.9 & 81.9 & 72.7 & 79.5 \\ 
    \multirow{-1}{*}{12$\times$224$^2$} & \textbf{12$\times$224$^2$} & 85.0 & 84.3 & 80.9 & 67.8 & 79.4  \\ 
    \multirow{1}{*}{Direct Tuned} & 16$\times$224$^2$ & 84.9 & 83.9 & 78.9 & 65.2 & 78.2 \\
    ~ & 20$\times$224$^2$ & 84.7 & 83.4 & 77.6 & 62.4 & 77.0 \\
    ~ & 24$\times$224$^2$ & 84.3 & 82.6 & 76.4 & 60.4 & 75.9 \\\cline{2-7}
    ~ & \textbf{Avg} & 84.7 & 83.6 & 79.4 & 67.2 & - \\ \cline{3-7}
    ~ & \textbf{Max} & 85.0 & 84.3 & 81.9 & 72.8 & - \\ 
    \Xhline{0.8pt}	
    ~ & 2$\times$224$^2$ & $\varnothing$ & $\varnothing$ & $\varnothing$ & 72.2 & 72.2 \\
    ~ & 4$\times$224$^2$ & $\varnothing$ & $\varnothing$ & 81.0 & 79.3 & 80.2 \\
    ~ & 8$\times$224$^2$ & $\varnothing$ & 84.4 & 82.8 & 80.3 & 82.5 \\
    ~ & 12$\times$224$^2$ & 85.4 & 85.2 & 83.3 & 79.9 & 83.5 \\
    (3072, 2048, 1024) & 16$\times$224$^2$ & 85.7 & 85.1 & 83.5 & 79.2 & 83.4 \\
    Flux-Multi Tuned & 20$\times$224$^2$ & 85.7 & 85.3 & 83.0 & 78.9 & 83.2 \\
    ~ & 24$\times$224$^2$ & 85.6 & 85.0 & 82.7 & 78.2 & 82.9 \\ \cline{2-7}
    ~ & \textbf{Avg} & 85.6 & 85.0 & 82.7 & 78.3 & - \\ \cline{3-7}
    ~ & \multirow{2}{*}{\textbf{Max}} & 85.7 & 85.3 & 83.5 & 80.3 & - \\  
    ~ & ~ & \darkGreen{$\uparrow$\textbf{0.7}}  & \darkGreen{$\uparrow$\textbf{1.0}}  & \darkGreen{$\uparrow$\textbf{1.6}} & \darkGreen{$\uparrow$\textbf{7.5}} & - \\ 
    \Xhline{1.0pt}	
    \end{tabular}
    }
    \vspace{-0.3cm}
    \caption{\textbf{Directly use Flux-Tuning with the previsou SOTA InternVideo2-S on K400.} Results are reported on K400 using 1clip$\times$1crop. It shows that Flux can be used as an advanced augmentation tool directly in supervised tuning scenario.}
    \label{tab:ablations_multi_k400_iv2}
    \vspace{-0.3cm}
\end{table}

\begin{figure}[t]
    \centering
    \includegraphics[width=\linewidth]{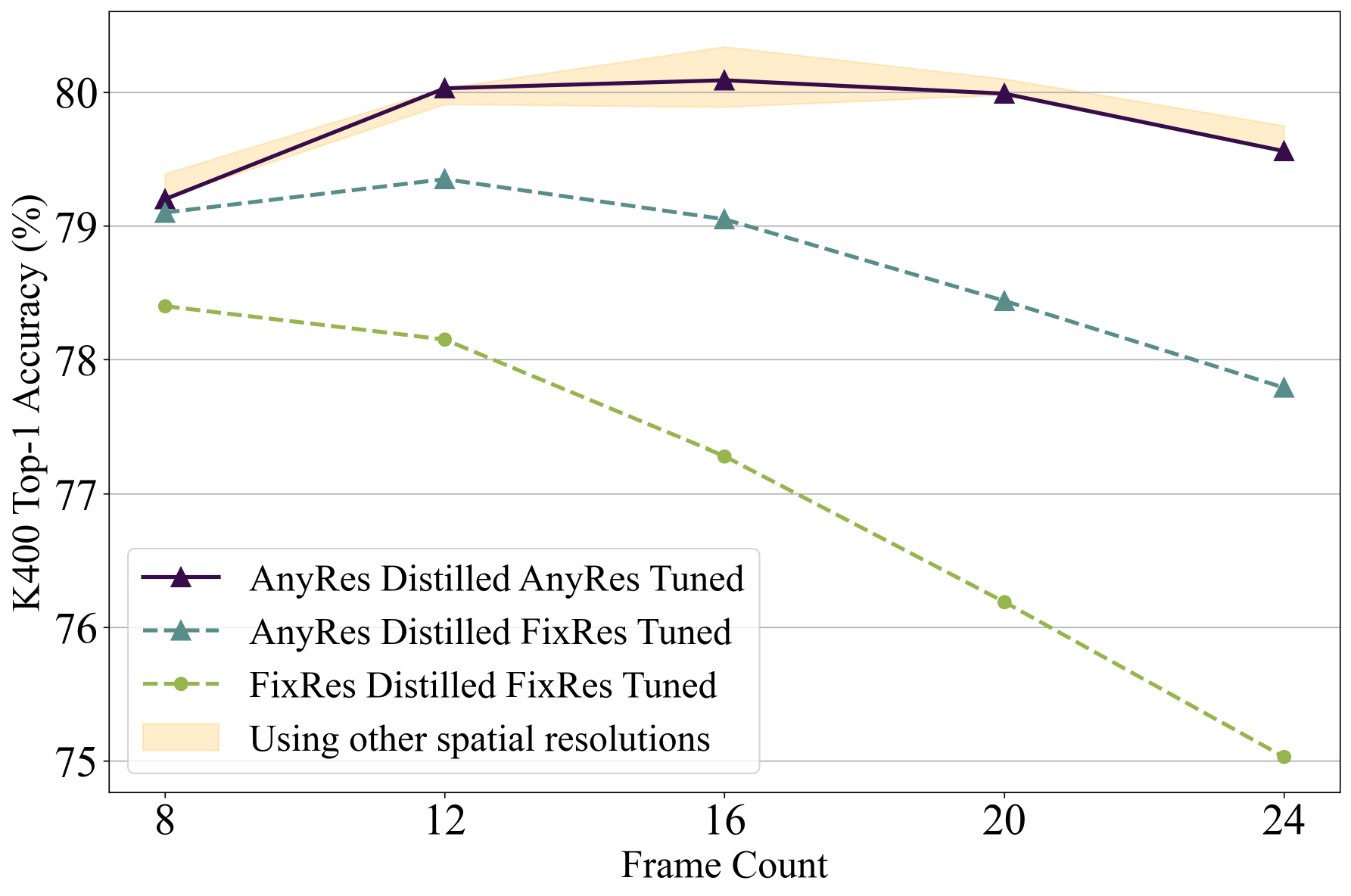}
    \vspace{-0.7cm}
    \caption{
        \textbf{Comparison between different training methods on K400 using a fixed number of 2048 tokens.}
        Note the three lines and all the points share similar training and inference costs. The shaded part shows results for the \textit{AnyRes Distilled AnyRes Tuned} model with spatial resolution in range (196, 252), while others use a fixed spatial resolution at 224.
    }
    \vspace{-0.1cm}
    \label{fig:anyres}
\end{figure}

\paragraph{Flux-Single Pre-training and Tuning.}
Figure \ref{fig:anyres} illustrates the performance of three models differing in their pre-training and tuning approaches. The first model, termed the \textit{FixRes Distilled FixRes Tuned} model, employs a fixed resolution of 8$\times$224$^2$ during both the pre-training stage and fine-tuning stage, the same as the original setting to get InternVideo2-S. We evaluate this tuned model using a fixed input token count of 2048 by our token-selector on the K400 dataset, which corresponds to the number of input tokens in the standard 8$\times$224$^2$ configuration, but with varied frame counts to assess the model's capability to leverage additional temporal information in fixed-number tokens for enhanced performance. Thus, \textbf{models trained and tuned with fixed sampling grid can hardly harvest costless performance gain with Token Optimization.}

In contrast, the \textit{AnyRes Distilled FixRes Tuned} model utilizes Flux-UMT-Single upon the original setting. Through Flux token selection, we maintain consistent pre-training computational costs by regulating a fixed quota of 2048 input tokens. The tuning method also remains unchanged. The diversified features from the teacher by the flexible sampling accumulate more valued representations for the student, gaining a +0.7\% improvement at the original 8$\times$224$^2$ test setting. Notably, the new model exhibits greater robustness in processing a fixed number of tokens from larger spatiotemporal resolution, achieving another +0.3\% gain with Token Optimization costlessly.

\newcommand{\performanceIncrease}[1]{
  (\textbf{\darkGreen{$\uparrow$\footnotesize{#1}}})
}

\newcommand{\performanceRelativeIncrease}[1]{
  (\textbf{\darkblue{$\uparrow$\footnotesize{#1}}})
}

\begin{table}[tp]
    \centering
    \setlength\tabcolsep{3pt}
    \resizebox{\linewidth}{!}{
        \begin{tabular}{l|l|l|l|l}
            \Xhline{1.0pt}	
            \textbf{Mask Type} & \textbf{K400} & \textbf{\textit{w/} TO} & \textbf{SSv2} & \textbf{\textit{w/} TO} \\
            \Xhline{1.0pt}	
            $\varnothing$ (single res) & 78.4 & 78.4 & 65.4 & 65.4 \\
            Random & 78.6 & 79.0\performanceIncrease{0.6} & 65.3 & 65.9\performanceIncrease{0.5} \\ \hline
            Tube & 78.8 & 80.0\performanceIncrease{1.6} & 65.7 & 66.7\performanceIncrease{1.3} \\
            Dynamic(L1) & 78.7 & 79.8\performanceIncrease{1.4} & 65.7 & 66.6\performanceIncrease{1.2} \\ 
            Dynamic(L2) & 78.8 & 80.0\performanceIncrease{1.6} & 65.8 & 66.7\performanceIncrease{1.3} \\ \hline
            Group-Dynamic(L2)-2 & 78.8 & 80.2\performanceIncrease{1.8}& 66.0 & 67.0\performanceIncrease{1.6} \\
            \rowcolor{blue!10} Group-Dynamic(L2)-4 & 79.2 & 80.3\performanceIncrease{1.9} & 66.3 & 67.3\performanceIncrease{1.9} \\ \Xhline{1.0pt}	
        \end{tabular}
    }
    \vspace{-0.3cm}
    \caption{\textbf{Ablation on using token selection strategies} We validate the effect of different token selection methods on Flux-Single-Pre-training and Tuning on ViT. A group size of 4 works well.}
    \label{tab:ablations_ofa}
    \vspace{-0.3cm}
\end{table}

\begin{table}[tp]
    \centering
    \setlength\tabcolsep{3.0pt}
    \resizebox{1.0\linewidth}{!}{
        \begin{tabular}{l|l|l|l|l}
            \Xhline{1.0pt}	
            \textbf{Method \& Arch} & \textbf{K400} & \textbf{\textit{w/} TO} & \textbf{SSv2} & \textbf{\textit{w/} TO} \\
            \Xhline{1.0pt}	
            {Vanilla + ViT} & 78.4 & 78.4 & 65.4 & 65.4 \\
            {Vanilla + FluxViT} & 79.3 & 79.6\performanceIncrease{1.2} & 66.0 & 66.4\performanceIncrease{1.0} \\ \hline
            {Flux-Single + ViT} & 79.2 & 80.3\performanceIncrease{1.9} & 66.3 & 67.3\performanceIncrease{1.9} \\  
            \multicolumn{5}{@{}l}{\darkGreen{\textit{With new positional embeddings}}} \\
            {\quad \textit{w/} RoPE} & 79.5 & 80.7\performanceRelativeIncrease{0.4} & 66.5 & 67.5\performanceRelativeIncrease{0.2} \\ 
            {\quad \textit{w/} GPE} & 79.4 & 80.5\performanceRelativeIncrease{0.2} & 66.4 & 67.4\performanceRelativeIncrease{0.1} \\ 
            {\quad \textit{w/} LPE} & 79.7 & 81.0\performanceRelativeIncrease{0.7}& 66.8 & 68.3\performanceRelativeIncrease{1.0} \\ 
            {\quad \textit{w/} GLPE} & 79.9 & 81.3\performanceRelativeIncrease{1.0} & 67.0 & 68.6\performanceRelativeIncrease{1.3} \\ 
            \multicolumn{5}{@{}l}{\darkGreen{\textit{With DPN}}} \\
            {\quad \textit{w/} DPN} & 79.8 & 81.2\performanceRelativeIncrease{0.9} & 66.9 & 68.4\performanceRelativeIncrease{1.1} \\ 
            \multicolumn{5}{@{}l}{\darkGreen{\textit{Combining the two modules}}} \\
            \rowcolor{blue!10} 
            {Flux-Single + FluxViT} & 80.5 & 81.7\performanceIncrease{3.3} & 67.6 & 69.3\performanceIncrease{3.9} \\ \hline 
            \rowcolor{blue!10} 
            {Flux-Multi + FluxViT} & 81.4 & 82.4\performanceIncrease{4.0} & 68.4 & 70.0\performanceIncrease{4.6} \\ \Xhline{1.0pt}	
        \end{tabular}
    }
    \vspace{-0.1cm}
    \caption{Results with 8$\times224^2$(and TO w/ 2048 tokens, which is the same token count as in 8$\times224^2$) on K400 and SSv2. \performanceIncrease{} for absolute improvement and \performanceRelativeIncrease{} for relative gain. Vanilla means PT and FT without Flux augmentation.}
    \label{tab:flux_single_ablation_refine}
    \vspace{-0.2cm}
\end{table}

\begin{table}[tp]
    \centering
    \setlength\tabcolsep{4pt}
    \resizebox{\linewidth}{!}{
        \begin{tabular}{l|cc|cc}
        \Xhline{1.0pt}	
        \multirow{2}{*}{\textbf{Settings}} & \multicolumn{2}{c|}{\textbf{8$\times$224$^2$}} & \multicolumn{2}{c}{\textbf{2048 + TO}} \\
        ~ & \textbf{K400} & \textbf{SSv2} & \textbf{K400} & \textbf{SSv2}  \\
        \Xhline{1.0pt}
        \rowcolor{blue!10}
        Baseline & \textbf{80.5} & 67.6 & \textbf{81.7} & \textbf{69.3} \\
        Change $T_{thres}$ to (2048, 6144) & 80.1 & 67.2 & 81.2 & 68.7 \\
        \textit{w/o} varied spatial resolution & \textbf{80.5} & \textbf{67.7} & 81.4 & 69.1 \\
        enlarge $F_{max}$ to 32 & 80.4 & 67.4 & 81.6 & 69.1 \\
        enlarge $Res_{max}$ to 336 & 80.2 & 67.3 & 81.4 & 69.0 \\
        \Xhline{1.0pt}	
        \end{tabular}
    }
    \vspace{-0.1cm}
    \caption{
    \textbf{Ablations on hyper-parameter on sampling designs for Flux-Single training with FluxViT.} Most settings regarding flexible sampling cause only minor influences except $T_{thres}$.
    }
    \label{tab:design_mm}
    \vspace{-0.2cm}
\end{table}

\begin{table}[t]
    \centering
    \setlength\tabcolsep{4.5pt}
    \resizebox{0.9\linewidth}{!}{
        \begin{tabular}{l|l|c|c|c}
        \Xhline{1.0pt}	
        \multicolumn{1}{c|}{\textbf{Pre-training}} & \multicolumn{1}{c|}{\textbf{Fine-tuning}} &\multicolumn{3}{c}{\textbf{Test Length}} \\ 
        \multicolumn{1}{c|}{\textbf{Length}} & \multicolumn{1}{c|}{\textbf{Length}} & {2048} & {1024} & {512} \\
        \Xhline{1.0pt}	
        $\varnothing$(\textit{w/o} Flux) & $\varnothing$ & 79.3 & 74.7 & 62.1 \\\hline
        Single & Single & 80.5 & 77.4 & 65.8 \\
        Single & Multi(\textit{w/} align) & 80.3 & 79.0 & 75.0 \\
        \shline
        Multi  & Single & 81.0 & 78.8 & 70.2 \\
        Multi  & Multi  & 80.9 & 79.1 & 73.3 \\
        \rowcolor{blue!10} 
        Multi  & Multi(\textit{w/} align)  & \textbf{81.4} & \textbf{80.3} & \textbf{76.6} \\ \Xhline{1.0pt}	
        \end{tabular}
    }
    \vspace{-0.1cm}
    \caption{\textbf{Ablation on using varied input lengths in training FluxViT on K400.} Results are tested with fixed 8$\times$224$^2$ input but varied test token lengths based on our selector.}
    \label{tab:ablations_multi_k400}
    \vspace{-0.1cm}
\end{table}

\begin{table}[t]
    \centering
    \setlength\tabcolsep{1pt}
    \resizebox{\linewidth}{!}{
        \begin{tabular}{l|l|r|r|cc}
        \Xhline{1.0pt}
            \textbf{Model} & \textbf{Extra Data} & \textbf{\#P(M)} & \textbf{GFLOPs}  & \multicolumn{2}{c}{\textbf{Top-1}} \\
            \Xhline{1.0pt}	
             TimeSformer-L~\cite{timesformer} & - & 121 & 2380$\times$3 & \multicolumn{2}{c}{80.7} \\
             VideoSwin-L ~\cite{video_swin} & IN-21K & 197 & 604$\times$12 & \multicolumn{2}{c}{83.1} \\
             VideoMAE-L ~\cite{videomae} & - & 305& 3958$\times$21 & \multicolumn{2}{c}{86.1} \\
             CoVeR-L ~\cite{cover} & JFT-3B+SMI & 431 & 5860$\times$3 & \multicolumn{2}{c}{87.1} \\
             \small{UniFormerV2-L} ~\cite{uniformerv2} &CLIP-400M+K710 & 354 & 12550$\times$6 & \multicolumn{2}{c}{90.0} \\
             UMT-L ~\cite{umt} & K710 & 431 & 5860$\times$3 & \multicolumn{2}{c}{90.6}  \\ \hline
             \small{VideoMAE2-H} ~\cite{videomaev2} & UnlabeledHybrid & 633 & 1192$\times$15 & \multicolumn{2}{c}{88.6} \\
             ViViT-H ~\cite{vivit} & JFT-300M & 654 & 3981$\times$12 & \multicolumn{2}{c}{84.9} \\
             MTV-H~\cite{mtv} & IN-21K+WTS-60M  & 1000+ & 6130$\times$12 & \multicolumn{2}{c}{89.9} \\
             CoCa-G ~\cite{coca} & JFT-3B+ALIGN-1.8B & 1000+ & N/A$\times$12 & \multicolumn{2}{c}{88.9} \\
             \hline
             MViTv1-B~\cite{mvit} & - & 37 & 70$\times$5 & \multicolumn{2}{c}{80.2} \\
             MViTv2-B~\cite{mvitv2} & - & 37 & 255$\times$5 & \multicolumn{2}{c}{81.2} \\
             ST-MAE-B~\cite{st_mae} & K600 & 87 & 180$\times$21 & \multicolumn{2}{c}{81.3} \\
             VideoMAE-B ~\cite{videomae} & - & 87 & 180$\times$15 & \multicolumn{2}{c}{81.5} \\
             VideoSwin-B~\cite{video_swin} & IN-21k & 88& 282$\times$12 & \multicolumn{2}{c}{82.7} \\
             UniFormer-B~\cite{uniformer} & IN-1k & 50 & 259$\times$12 & \multicolumn{2}{c}{83.0} \\
             UMT-B~\cite{umt} & K710 & 87 & 180$\times$12 & \multicolumn{2}{c}{87.4} \\
             InternVideo2-B~\cite{iv2} & K710+MASH & 96 & 440$\times$12 & \multicolumn{2}{c}{88.4} \\ \hline
             \rowcolor{blue!10} ~ & ~ & ~ & 440$\times$12 & 88.7 & \textbf{\darkblue{89.4}}  \\
             \rowcolor{blue!10} \multirow{-2}{*}{\textbf{FluxViT-B$_{e200}$}} & \multirow{-2}{*}{-} &  \multirow{-2}{*}{97} & 49$\times$12 & 84.0 & \textbf{\darkblue{86.7}} \\ \hline
             \rowcolor{blue!10} 
             ~ & & ~ & 440$\times$12 & 89.6 & \textbf{\darkblue{90.0}} \\
             \rowcolor{blue!10} 
             ~ & & ~ & 255$\times$12 & 89.3 & \textbf{\darkblue{89.7}} \\
             \rowcolor{blue!10} 
             ~ & & ~ & 108$\times$12 & 87.3 & \textbf{\darkblue{88.9}} \\
             \rowcolor{blue!10} 
             \multirow{-4}{*}{\textbf{FluxViT-B$_{e100}$}} &\multirow{-4}{*}{K710+MASH} & \multirow{-4}{*}{97} & 49$\times$12 & 84.7 & \textbf{\darkblue{87.4}} \\
             \Xhline{0.8pt}
             UniFormer-S~\cite{uniformer} & IN-1k & 21&  42$\times$4 & \multicolumn{2}{c}{80.8} \\
             MViTv2-S~\cite{mvitv2} & - & 35 &  64$\times$5 & \multicolumn{2}{c}{81.0} \\
             VideoMAE-S ~\cite{videomae} & - & 22 & 57$\times$15 & \multicolumn{2}{c}{79.0} \\
             VideoMAE2-S ~\cite{videomaev2} & - & 22 & 57$\times$15 & \multicolumn{2}{c}{83.7} \\
             InternVideo2-S~\cite{iv2} & K710+MASH & 23 & 154$\times$12 & \multicolumn{2}{c}{85.8} \\ \hline
             \rowcolor{blue!10} ~ & ~ & ~ & 154$\times$12 & 86.4 & \textbf{\darkblue{87.3}}  \\ 
             \rowcolor{blue!10} \multirow{-2}{*}{\textbf{FluxViT-S$_{e200}$}} &  \multirow{-2}{*}{-} &  \multirow{-2}{*}{24} & 13$\times$12 & 79.7 & \textbf{\darkblue{84.0}}  \\ \hline
             \rowcolor{blue!10} ~ & & ~ & 154$\times$12 & 87.7 & \textbf{\darkblue{88.0}}  \\
             \rowcolor{blue!10} ~ & & ~ &  83$\times$12 & 87.3 & \textbf{\darkblue{87.7}} \\
             \rowcolor{blue!10} ~ & & ~ &  32$\times$12 & 84.7 & \textbf{\darkblue{86.6}} \\
             \rowcolor{blue!10} \multirow{-4}{*}{\textbf{FluxViT-S$_{e100}$}} & \multirow{-4}{*}{K710+MASH} & \multirow{-4}{*}{24} &  13$\times$12 & 80.1 & \textbf{\darkblue{84.7}} \\
        \Xhline{1.0pt}	
        \end{tabular}
    }
    \vspace{-0.2cm}
    \caption{\textbf{Comparison with the state-of-the-art methods with on scene-related Kinetics-400.} \#P is short for the number of parameters. The \darkblue{blue} values of \ModelName{} show results using larger spatiotemporal resolutions but keeping fixed input token count to 3072, 2048, 1024, and 512 respectively, corresponding to the four GFLOPs listed. SMI is short for the train set of SSv2, MiT and ImageNet and MASH for MiT, ANet, SSv2 and HACS.
    }
    \vspace{-0.1cm}
    \label{results_k400}
\end{table}

We further test tuning the any-res trained model using Flux-FT with a fixed input token number of 2048 on K400, as termed \textit{AnyRes Distilled AnyRes Tuned} in the figure. As shown, this tuning method can harvest a +1.9\% top-1 accuracy gain on K400 compared with the normal distilled and normal tuned method while nearly bringing no computation increment throughout pre-training and fine-tuning.

Table \ref{tab:ablations_multi_k400_iv2} further shows that only combining Flux-Tuning and Token Optimization on a competitive model pre-trained with standard setting can achieve non-trivial gain, highlighting that the sampling and selection module in Flux can effectively serve as novel augmentation tools.

\paragraph{Token Selection Module.}
We test four token selection methods for Flux-Single, including random, tube, dynamic (as also directly used in \cite{Hwang2022EVERESTEM}) and our group-dynamic method. We further ablate the token dynamics measurement and the sparse group size. Table \ref{tab:ablations_ofa} shows that our group-dynamic strategy with token dynamics measured by L2-distance of tokens within adjacent frames in sparsely divided four groups is the most robust among the token-selection methods. More advanced token selectors like Token Merging~\cite{tome} and Vid-TLDR~\cite{vidtldr} can be utilized for better results but with increased complexity, cost, and tedious hyperparameters. We provide experiments using Vid-TLDR in the supplement material and advocate the current selector due to its simplicity.

\paragraph{FluxViT Modules.}
Table \ref{tab:flux_single_ablation_refine} analyzes the impact of our proposed plug-in modules. The inclusion of Global-Local positional embedding can strengthen the model's robustness when processing sparse tokens derived from larger spatiotemporal resolutions. We compare our GLPE with ROPE~\cite{su2024roformer} within our Flux scheme and show better results with LPE illustrated in the Method section. Furthermore, adding a dual patch norm module optimizes the performance gains in TO. Combining the two modules achieves +3.3\% and +3.9\% on K400 and SSv2 respectively with TO.

\begin{table}[t]
    \centering
    \setlength\tabcolsep{2pt}
    \resizebox{\linewidth}{!}{
        \begin{tabular}{l|l|r|cc|cc}
        \Xhline{1.0pt}
            \textbf{Model} & \textbf{Extra Data} & \textbf{GFLOPs}  & \multicolumn{2}{c|}{\textbf{Top-1}} & \multicolumn{2}{c}{\textbf{Top-5}} \\
            \Xhline{1.0pt}	
             TimeSformer-L~\cite{timesformer} & IN-21k & 2380$\times$3 & \multicolumn{2}{c|}{62.3} & \multicolumn{2}{c}{-} \\
             MViTv1-B~\cite{mvit} & K400 & 455$\times$3 & \multicolumn{2}{c|}{67.7} & \multicolumn{2}{c}{70.9} \\
             MViTv2-B~\cite{mvitv2} & K400 & 255$\times$3 & \multicolumn{2}{c|}{70.5} & \multicolumn{2}{c}{92.7} \\
             VideoMAE-B ~\cite{videomae} & K400 & 180$\times$6 & \multicolumn{2}{c|}{69.7} & \multicolumn{2}{c}{92.3} \\
             VideoMAE-L ~\cite{videomae} & K400 & 596$\times$6 & \multicolumn{2}{c|}{74.0} & \multicolumn{2}{c}{94.6} \\
             UniFormerV2-B ~\cite{uniformerv2} & CLIP-400M & 375$\times$3 & \multicolumn{2}{c|}{70.7} & \multicolumn{2}{c}{93.2} \\
             UniFormerV2-L ~\cite{uniformerv2} & CLIP-400M & 1718$\times$3 & \multicolumn{2}{c|}{73.0} & \multicolumn{2}{c}{94.5} \\
             UMT-B~\cite{umt} & K710 & 180$\times$6 & \multicolumn{2}{c|}{70.8} & \multicolumn{2}{c}{92.4} \\
             InternVideo2-B~\cite{iv2} & K710+MASH & 253$\times$6 & \multicolumn{2}{c|}{73.5} & \multicolumn{2}{c}{94.4} \\
             \rowcolor{blue!10} 
             ~ & & 440$\times$6 & 75.3 & \textbf{\darkblue{75.6}} & 95.1 & \textbf{\darkblue{95.1}} \\
             \rowcolor{blue!10} 
             ~ & & 255$\times$6 & 75.1 & \textbf{\darkblue{75.5}} & 94.9 & \textbf{\darkblue{95.1}} \\
             \rowcolor{blue!10} 
             ~ & & 108$\times$6 & 72.0 & \textbf{\darkblue{75.1}} & 93.3 & \textbf{\darkblue{94.8}} \\
             \rowcolor{blue!10}
             \multirow{-4}{*}{\textbf{FluxViT-B}} & \multirow{-4}{*}{K710+MASH} & 49$\times$6 & 56.8 & \textbf{\darkblue{73.9}}  & 84.8 & \textbf{\darkblue{94.4}} \\
             \Xhline{0.8pt}
             UniFormer-S~\cite{uniformer} & IN-1K & 42$\times$3 & \multicolumn{2}{c|}{67.7} & \multicolumn{2}{c}{91.4} \\
             VideoMAE-S ~\cite{videomae} & K600 & 57$\times$6 & \multicolumn{2}{c|}{66.8} & \multicolumn{2}{c}{90.3} \\
             InternVideo2-S~\cite{iv2} & K710+MASH & 83$\times$6 & \multicolumn{2}{c|}{71.5} & \multicolumn{2}{c}{93.4} \\
             \rowcolor{blue!10} ~ & & 154$\times$6 & 73.4 & \textbf{\darkblue{73.8}} &  94.1 & \textbf{\darkblue{94.1}} \\
             \rowcolor{blue!10} ~ & &  83$\times$6 & 72.9 & \textbf{\darkblue{73.4}} &  94.0 & \textbf{\darkblue{94.1}} \\
             \rowcolor{blue!10} ~ & &  32$\times$6 & 70.0 & \textbf{\darkblue{72.5}} &  93.4 & \textbf{\darkblue{93.8}} \\
             \rowcolor{blue!10} \multirow{-4}{*}{\textbf{FluxViT-S}} & \multirow{-4}{*}{K710+MASH} & 13$\times$6 & 55.3 & \textbf{\darkblue{70.9}} & 83.7 & \textbf{\darkblue{93.1}} \\
        \Xhline{1.0pt}	
        \end{tabular}
    }
    \vspace{-0.2cm}
    \caption{\textbf{Comparison with the state-of-the-art methods with on motion-intensive SSv2.} Our model achieves far better results.
    }
    \vspace{-0.1cm}
    \label{tab:results_ssv2}
\end{table}

\paragraph{Hyper-parameters of flexi-sampling}
Table \ref{tab:design_mm} gives ablation studies on the results of different hyper-parameters regarding sampling. Most settings regarding flexible sampling cause only minor influences except $T_{thres}$. A proper scale of the flexible space is the most powerful.

\paragraph{Flux-Multi training.}
We compare our proposed method of co-training with three input numbers(2048, 1024, 512) in both pretraining and tuning in Table \ref{tab:ablations_multi_k400} and also validate the tuning effects on well-pretrained InternVideo2-S in table \ref{tab:ablations_multi_k400_iv2}(3072, 2048, 1024). As shown, pre-training with multiple token numbers can not only boost performance under standard settings but also increase model consistency in input token sparsity. Tuning with different token numbers with a smoothed-L1-loss based self-distillation mechanism can further enhance such alignment consistently. As the multi-token-number training brings additional computation, we report the efficiency here \textbf{(i)} Time: For ablation setting, Flux-Single-UMT takes 15.5 hours using 32 A100 and Multi takes 20.3. \textbf{(ii)} GPU memory: with a per-gpu batch size of 32, Flux-Single takes 44GB GPU memory while Multi takes 70GB. Flux-Multi achieves far better results at varied token counts, with acceptable overhead, as it best leverages the heavy teacher-forward process.

\subsection{Single Modality Results}
We scale up the training data using the K-MASH\cite{iv2} dataset of 1.1 million samples, aligning with the data employed in the Internvideo2-Distilled series models, including K710, SSv2\cite{goyal2017something}, ANet\cite{activitynet}, HACS\cite{hacs}, and MiT\cite{mit}. We train the FluxViT model using a total batch size of 2048 for 100 epochs. FluxViT's performance is validated using scene-based K400, motion-intensive SSv2, and long-term COIN with Token Optimization. 

\paragraph{K400.}
Table \ref{results_k400} reports our results compared with the previous solutions on K400. We set 3072, 2048, and 1024 as our token numbers in Flux-Multi-Tuning. For our \ModelName{}-S, we report \textbf{88.0\%}(+2.2\%) on K400 compared with InternVideo2-S, which is the previous SOTA solution, while still achieving \textbf{84.7\%} with nearly the same computation as the lightweight network UniFormer~\cite{uniformer} but with competitive accuracy. For fair comparison without K-MASH, we pretrain FluxViT using only K400 in Flux-UMT, which also shows much better results.

\paragraph{SSv2.}
Table \ref{tab:results_ssv2} shows our model's performance in dealing with motion-intensive video understanding tasks. Our FluxViT-S and FluxViT-B models set new state-of-the-art performance on SSv2 with either standard or limited computation cost. For FluxViT-B, we achieve 75.6\% with standard computation while still achieving 75.1\% with only 25\% cost of the standard setting. Previous work ~\cite{frameflexiblenet} instead gets a huge performance drop using 2 frames, which corresponds to the least computation listed.

\begin{table}[t]
    \centering
    \setlength\tabcolsep{2pt}
    \resizebox{0.9\linewidth}{!}{
        \begin{tabular}{l|c|l|cc}
        \Xhline{1.0pt}
            \textbf{Model} & \textit{\textbf{e2e}} & \textbf{BackBone} & \multicolumn{2}{c}{\textbf{Top-1}} \\  
            \Xhline{1.0pt}	
            Distant Supervision~\cite{distant} & \xmark & TimeSformer & \multicolumn{2}{c}{90.0} \\
            ViS4mer~\cite{vis4mer} & \xmark & Swin-B & \multicolumn{2}{c}{88.4} \\
            \hline
            Turbo$_{f32}$~\cite{turbo} & \cmark & VideoMAE-B & \multicolumn{2}{c}{87.5} \\
            VideoMamba$_{f64}$~\cite{videomamba}  & \cmark & VideoMamba-S & \multicolumn{2}{c}{88.7} \\
            VideoMamba$_{f64}$~\cite{videomamba}  & \cmark & VideoMamba-M & \multicolumn{2}{c}{90.4} \\
            InternVideo2$_{f12}$~\cite{iv2} & \cmark & InternVideo2-S & \multicolumn{2}{c}{90.0} \\
            \hline
            \gray{MA-LMM ~\cite{mallm}} & \gray{\cmark} & \gray{MLLM} & \multicolumn{2}{c}{\gray{93.2}} \\
            \gray{HERMES ~\cite{hermes}} & \gray{\cmark} & \gray{MLLM} & \multicolumn{2}{c}{\gray{93.5}} \\
            \Xhline{0.8pt}	
            \rowcolor{blue!10} \textbf{FluxViT}$_{3072}$ & \cmark & FluxViT-S & 91.8 & \textbf{\darkblue{92.1}} \\ 
            \rowcolor{blue!10} \textbf{FluxViT}$_{2048}$ & \cmark & FluxViT-S & 91.5 & \textbf{\darkblue{91.9}} \\ 
            \rowcolor{blue!10} \textbf{FluxViT}$_{1024}$ & \cmark & FluxViT-S & 89.8 & \textbf{\darkblue{91.0}} \\ \hline
            \rowcolor{blue!10} \textbf{FluxViT}$_{3072}$ & \cmark & FluxViT-B & 93.9 & \textbf{\darkblue{94.1}} \\ 
            \rowcolor{blue!10} \textbf{FluxViT}$_{2048}$ & \cmark & FluxViT-B & 93.7 & \textbf{\darkblue{93.9}} \\ 
            \rowcolor{blue!10} \textbf{FluxViT}$_{1024}$ & \cmark & FluxViT-B & 92.5 & \textbf{\darkblue{93.2}} \\
            \Xhline{1.0pt}
        \end{tabular}
    }
    \vspace{-0.2cm}
    \caption{\textbf{Comparison with the state-of-the-art on long-form video classification COIN dataset.} We report the results based on our preset token number, with the left line using unmasked 12, 8, 4 frames, and 224 spatial resolution while the \darkblue{blue} values show results that can be achieved using more informative tokens.
    }
    \label{results_coin}
    \vspace{-0.2cm}
\end{table}  

\paragraph{COIN.}
Table \ref{results_coin} shows our model's performance in dealing with long-form video understanding tasks. Our FluxViT-S and FluxViT-B models set new state-of-the-art performance on COIN with either standard or limited computation cost. 

\begin{table}[tp]
    \centering
    \small
    \setlength\tabcolsep{2pt}
    \resizebox{\linewidth}{!}{
    \begin{tabular}{l|ccccc}
        \Xhline{1.0pt}
        {\bf Model} & \textbf{MSR} & \textbf{DDM} & \textbf{ANet} & \textbf{LSMDC} & \textbf{MSVD}  \\
        \Xhline{1.0pt}	
        Internvideo2-S$_{2048}$~\cite{iv2} & 35.6 & 33.7 & 34.5 & 14.7 & 41.8 \\
        \hline
        Frozen-B~\cite{bain2021frozen}  & 18.7 & 20.2 & - & - & -\\
        VIOLET-B ~\cite{fu2021violet}  & 25.9 & 23.5 & - & - & - \\
        Singularity-B ~\cite{lei2022revealing} & 34.0 & 37.1 & 30.6 & - & - \\
        OmniVL-B ~\cite{wang2022omnivl}  & 34.6 & 33.3 & - & - & -\\ 
        CLIP4Clip-B ~\cite{luo2022clip4clip} & 30.6 & - & - & 13.6 & 36.2 \\
        UMT-B~\cite{umt} & 35.2 & 41.2 & 35.5 & 19.1 & 42.3 \\
        Internvideo2-B$_{2048}$~\cite{iv2} & 40.3 & 40.3 & 41.5 & 18.7 & 49.1 \\
        \hline
        VINDLU-L ~\cite{Cheng2022VindLUAR} & 32.0 & 36.9 & 30.9 & - & - \\
        InternVideo-L ~\cite{Wang2022InternVideoGV} & 40.7 & 31.5 & 30.7 & 17.6 & 43.4 \\ 
        UMT-L ~\cite{umt} & 40.7 & 48.6 & 41.9 & 24.9 & 49.0 \\
        ViClip-L ~\cite{internvid} & 42.4 & 18.4 & 15.1 & 20.1 & 49.1 \\
        InternVideo2-L ~\cite{iv2} & 42.1 & 42.8 & 43.6 & 21.4 & - \\
        LanguageBind-L ~\cite{zhu2023languagebind} & 42.8 & 39.7 & 38.4 & - & 54.1 \\
        LanguageBind-H ~\cite{zhu2023languagebind} & 44.8 & 39.9 & 41.0 & - & 53.9 \\
        VideoCoCa-G ~\cite{videococa} & 34.3 & - & 34.5 & - & - \\
        VideoPrism-G ~\cite{videococa} & 39.7 & - & 52.7 & - & - \\
        VAST-G ~\cite{vast} & 49.3 & 55.5 & - & - & - \\
        \Xhline{0.8pt}
        \rowcolor{blue!10}
        ~ & 44.4 & 48.3 & 52.4 & 20.8 & 49.4 \\
        \rowcolor{blue!10}
        \multirow{-2}{*}{\textbf{FluxViT-S$_{2048}$}} & \textbf{45.0} & \textbf{49.3} & \textbf{52.4} & \textbf{22.4} & \textbf{49.7} \\ 
        \cline{2-6}
        \rowcolor{blue!10}
        ~ & 42.2 & 45.4 & 47.2 & 18.7 & 48.1 \\
        \rowcolor{blue!10}
        \multirow{-2}{*}{\textbf{FluxViT-S$_{1024}$}} & \textbf{44.5} & \textbf{49.0} & \textbf{50.3} & \textbf{20.5} & \textbf{48.5} \\ 
        \cline{2-6}
        \rowcolor{blue!10}
        ~ & 36.8 & 38.5 & 38.2 & 17.7 & 45.5 \\
        \rowcolor{blue!10}
        \multirow{-2}{*}{\textbf{FluxViT-S$_{512}$}} & \textbf{40.5} & \textbf{45.8} & \textbf{44.7} & \textbf{19.0} & \textbf{46.9} \\ 
        \hline
        \rowcolor{blue!10}
        ~ & 49.8 & 52.2 & 56.6 & 23.7 & 53.8 \\
        \rowcolor{blue!10}
        \multirow{-2}{*}{\textbf{FluxViT-B$_{2048}$}} & \textbf{49.9} & \textbf{53.5} & \textbf{56.7} & \textbf{25.4} & \textbf{54.2} \\ 
        \cline{2-6}
        \rowcolor{blue!10}
        ~ & 48.0 & 48.8 & 51.8 & 22.6  & 52.8 \\
        \rowcolor{blue!10}
        \multirow{-2}{*}{\textbf{FluxViT-B$_{1024}$}} & \textbf{49.1} & \textbf{53.0} & \textbf{54.8} & \textbf{24.1} & \textbf{53.4} \\ 
        \cline{2-6}
        \rowcolor{blue!10}
        ~ & 42.6 & 42.9 & 42.8 & 20.1 & 50.7 \\
        \rowcolor{blue!10}
        \multirow{-2}{*}{\textbf{FluxViT-B$_{512}$}} & \textbf{47.2} & \textbf{49.8} & \textbf{50.3} & \textbf{22.8} & \textbf{52.1} \\
        \Xhline{1.0pt}
    \end{tabular}
    }
    \vspace{-0.2cm}
    \caption{\textbf{Zero-shot text-to-video retrieval on MSRVTT (``MSR''), DiDeMo (``DDM''), AcitivityNet (``ANet''), LSMDC, and MSVD.}
    We only report the R@1 accuracy. The upper line regarding FluxViT shows results with non-masked 8$\times$224$^2$, 4$\times$224$^2$ and 2$\times$224$^2$ input setting as indicated by the token count while each lower \textbf{bold} line shows results further using more informative tokens. We employ Dual Softmax Loss for the results.
    }
    \label{tab:retrieval_zs}
    \vspace{-0.3cm}
\end{table}

\subsection{Multi Modality Retrieval Results}
We use the pre-trained FluxViT with the CLIP~\cite{clip} framework and our Flux method to train a video clip model using 27M corpus, far lower than most baseline models, including 25M coarse level caption data: Webvid10M~\cite{bain2021frozen}, CC3M~\cite{cc3m}, COCO~\cite{coco}, VG~\cite{vg}, SBU~\cite{sbu}, CC12M~\cite{cc12m} and 2M high-quality data: S-MiT~\cite{smit}, InternVid-2M-Recap~\cite{internvid}. We leverage MobileClip-B~\cite{vasu2024mobileclip} as the text encoder and only use the vanilla VTC loss, which is short for Video-Text Contrastive loss in training. We only unfreeze the ViT projector for the first stage using the 25M coarse caption data for 3 epochs. Then we unfreeze all the modules for the second stage using the 2M high-quality data for one epoch to fully boost the ViT's capacity. By default, we train the clip model with a batch size of 4096. Also, we train the clip model with three input token numbers, 2048, 1024, and 512, and utilize smoothed-L1-loss on the final aggregated vision features to perform self-distillation and compute contrastive loss for each vision feature of the three numbers. Table \ref{tab:retrieval_zs} indicates that the clip model trained with our \ModelName{} method outperforms the Internvideo2-Series model with a large margin and further surpasses most of the top-performing models in Large or even Giant scale. Results on zero-shot action recognition results are included in Appendix for full validation of FluxViT-CLIP.

\begin{table}[tp]
    \centering
    \setlength\tabcolsep{2pt}
    \resizebox{\linewidth}{!}{
        \begin{tabular}{l|c|c|c|c}
            \Xhline{1.0pt}
            \textbf{Encoder} & \textbf{\#Tokens} & \textbf{\textit{w/} TO} & \textbf{MVbench} & \textbf{Dream1k-F1} \\ \Xhline{1.0pt}
            Clip-L~\cite{clip} & 8$\times$256 & \textit{\xmark} & 45.6 & 28.4 \\ 
            SigLIP$_{336}$-L~\cite{siglip} & 8$\times$576 & \textit{\xmark} & 46.7 & 29.2 \\ 
            InternVideo2-L~\cite{iv2} & 8$\times$224 & \textit{\xmark} & 47.0 & 28.7 \\ \hline 
            SigLIP$_{336}$-L~\cite{siglip} & 4$\times$576 & \textit{\xmark} & 44.5 & 25.4 \\
            UMT-L~\cite{umt} & 4$\times$256 & \textit{\xmark} & 45.0 & 24.6 \\ 
            \Xhline{0.8pt}
            \rowcolor{blue!10}~ & 8$\times$256 & & 48.3 & 29.0 \\
            \rowcolor{blue!10}~ & 4$\times$256 & & 46.9 & 27.9 \\
            \rowcolor{blue!10}\multirow{-3}{*}{\textbf{FluxViT-L}} & 2$\times$256 & \multirow{-3}{*}{\textit{\xmark}} & 46.0 & 25.6 \\ \hline
            \rowcolor{blue!10}~ & 2048 & & 49.0\performanceIncrease{0.7} & 29.5\performanceIncrease{0.5} \\
            \rowcolor{blue!10}~ & 1024 & & 47.7\performanceIncrease{0.8} & 28.5\performanceIncrease{0.6} \\
            \rowcolor{blue!10}\multirow{-3}{*}{\textbf{FluxViT-L}} & 512 & \multirow{-3}{*}{\textit{\cmark}} & 47.6\performanceIncrease{1.6} & 27.5\performanceIncrease{1.9} \\ \Xhline{1.0pt}
        \end{tabular}
    }
    \vspace{-0.3cm}
    \caption{\textbf{Results on Chat-Centric benchmarks MVbench (General perception) and Dream1k(Fine-grained caption).} Models are trained in a multimodal `linear prob' setting where both the Encoder and the LLM are frozen. \#Tokens for the number of visual tokens by the vision encoder.}
    \label{tab:flux_chat}
    \vspace{-0.1cm}
\end{table}

\subsection{Chat-Centric Evaluation Results}
We scale the pre-trained model size to FluxViT-Large and evaluate its performance under the VideoChat framework. We employ a single-layer MLP projector between ViT and an LLM(Qwen2-7B\cite{wang2024qwen2}) and train only the projector to ensure an unbiased evaluation of the ViT's performance, termed `linear probe' setting in multimodal scenarios. Such a probing strategy is widely used in the Stage-1 training of multi-modal chat models. We utilize a large-scaled train-set including LLava-558K\cite{llava}, S-MiT, 700k filtered subset of WebVid-10M, VidLN\cite{vidln}, and SSv2-open-ended. We then compare the models' performance on two benchmarks: MVbench, which assesses general spatiotemporal perception capabilities, and Dream1k, which evaluates the model's fine-grained ability to generate detailed captions. The other models tested include CLIP-L, SigLIP-L, and UMT-L, which are widely used in Video MLLMs. Given that ViT and LLM remain frozen, we maintain the original input settings as their pretraining phases to ensure comparability. We will scale the corpus size and fully tune the chat model for full performance in future work.
\section{Conclusion and Future work}
\label{sec:conclusion}
We present a token optimization process to maximize information in limited tokens within any budget. It integrates seamlessly with mainstream frameworks, offering a simple and scalable method to enhance real-world applications and facilitate future video foundation models. Future work could explore more advanced token selection methods for improvement, as the group setting can ensure full video coverage but not fully resist consistent camera motions. We test advanced Vid-TLDR~\cite{vidtldr} in Appendix, but with increased cost, tedious hyper-parameter tuning, and unstable improvement. For all current tasks tested, including fine-grained captioning, our method is still simple and effective.


\appendix

\section{More Experiments}
\label{sec:moreexp}

\begin{figure}[tp]
    \centering
    \includegraphics[width=\linewidth, keepaspectratio]{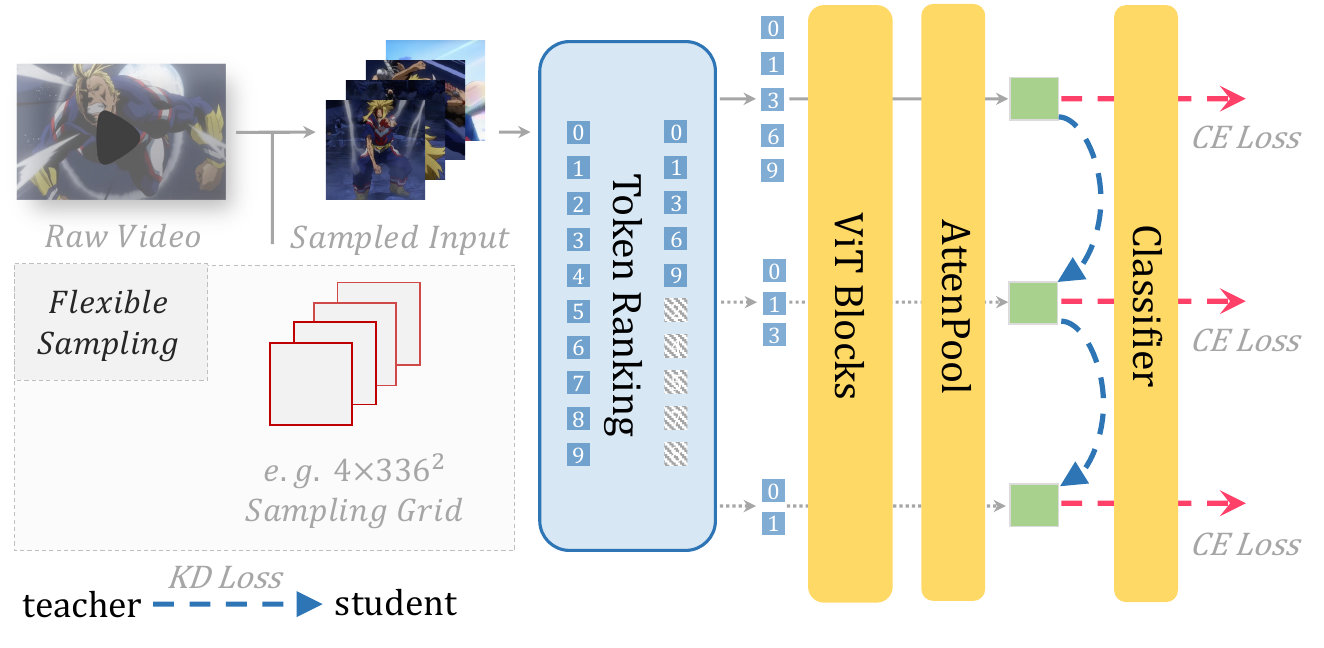}
    \vspace{-0.6cm}
    \caption{
    \textbf{Overview of Flux-Multi Tuning.}  
    }
    \vspace{-0.2cm}
    \label{fig:framework_supp_multi}
\end{figure}

\begin{figure}[tp]
    \centering
    \includegraphics[width=\linewidth, keepaspectratio]{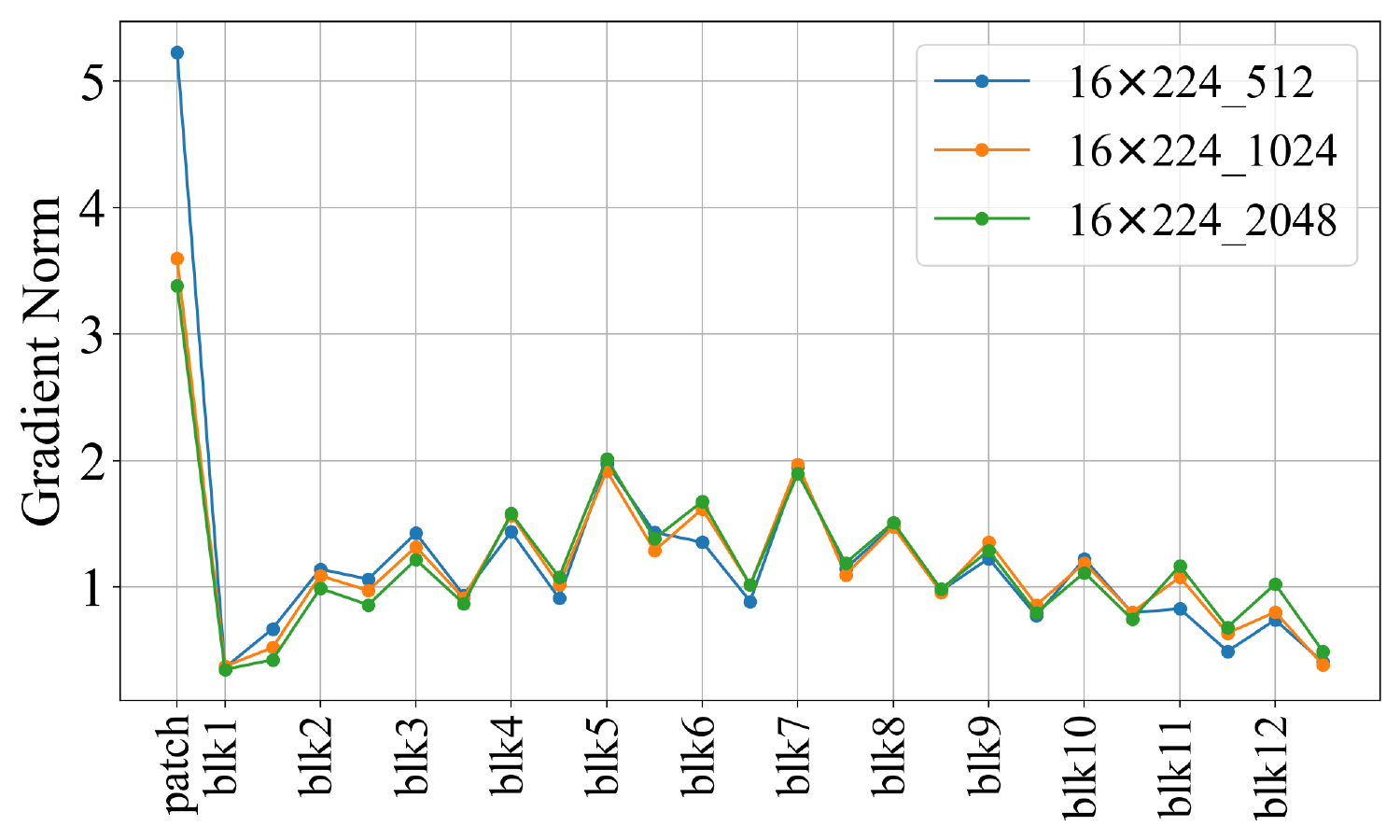}
    \vspace{-0.6cm}
    \caption{
    \textbf{Gradient norms of main projector modules of Flux-Multi trained InternVideo2 on K400.} We report the L2 gradient norm using bs=32.
    }
    \vspace{-0.2cm}
    \label{fig:unstable}
\end{figure}

\subsection{More ablation studies}

We here provide more ablation studies on Flux, including analyzing Flux's training stability with gradient norm, full results using different spatiotemporal resolutions and a corresponding heuristic TO validation strategy, and convergence analysis and experimenting with possible token merging methods in Flux. This will provide a more in-depth analysis of the whole Flux method. 


\begin{figure}[tp]
    \centering
    \includegraphics[width=\linewidth, keepaspectratio]{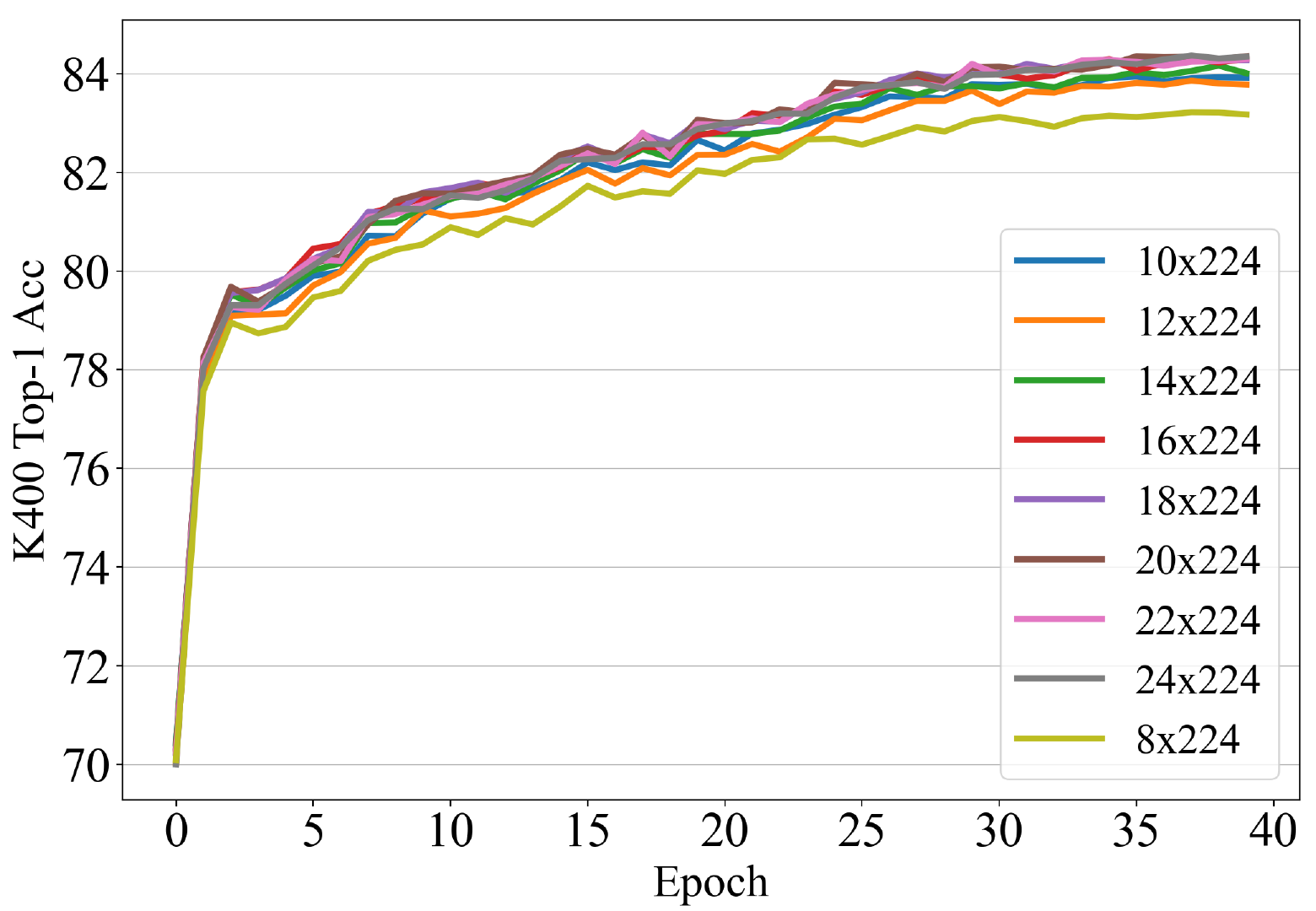}
    \vspace{-0.6cm}
    \caption{
    \textbf{Convergence analysis of Flux-Single tuning using 3072 tokens but different frame counts directly on K400.}  
    }
    \vspace{-0.2cm}
    \label{fig:converg_analysis}
\end{figure}

\begin{figure}[tp]
    \centering
    \includegraphics[width=\linewidth, keepaspectratio]{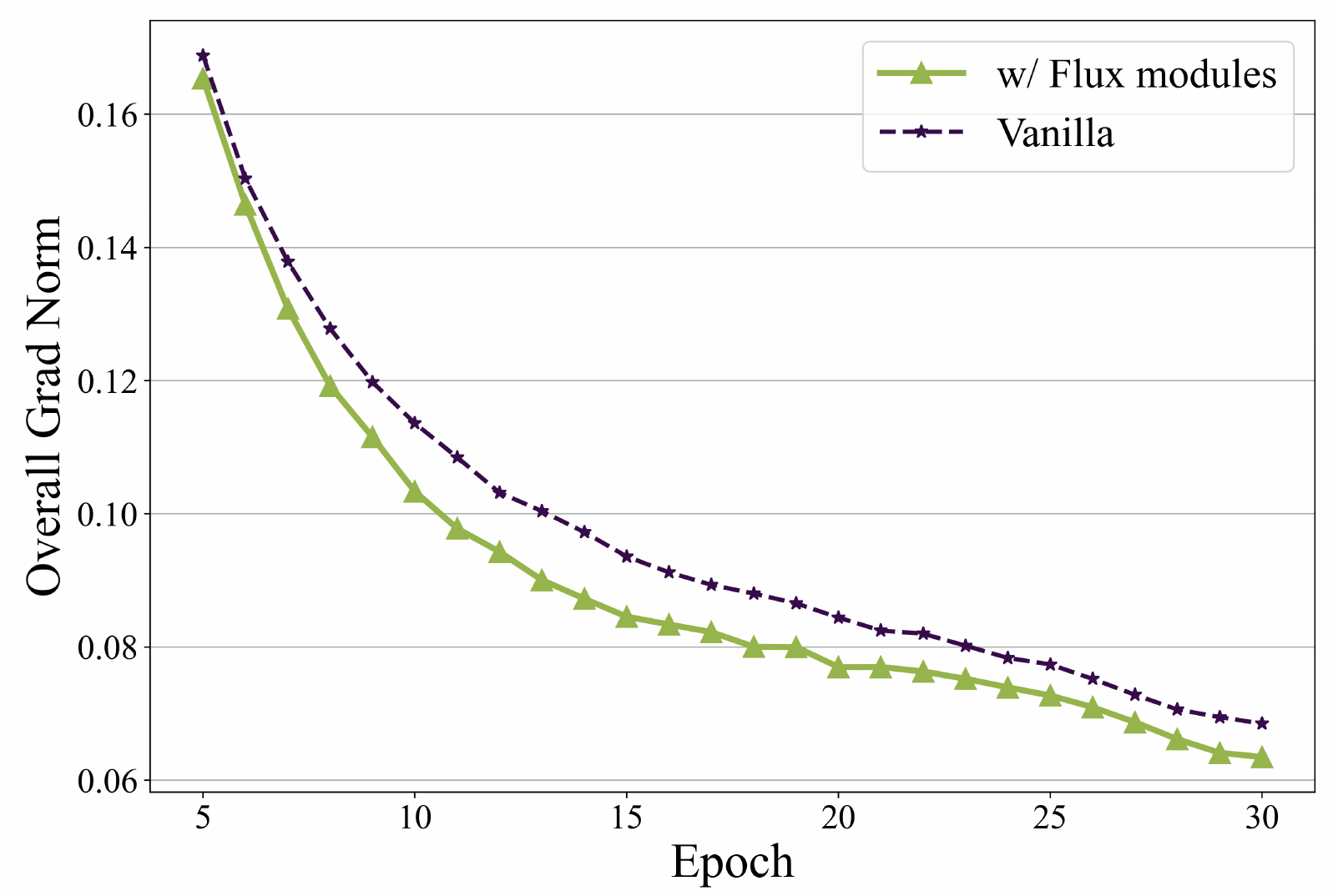}
    \vspace{-0.6cm}
    \caption{
    \textbf{Overall gradient norm trend during Flux-UMT per-training.} We report the overall training dynamics with our ablation setting. The FluxViT modules can lower the overall norm. 
    }
    \vspace{-0.2cm}
    \label{fig:overall_gnorm}
\end{figure}

\begin{table}[t]
    \centering
    \setlength\tabcolsep{1.2pt}
    \resizebox{0.65\linewidth}{!}{
    \begin{tabular}{l|c|c|c|c|c|c}
    \Xhline{1.0pt}	
    \multirow{2}{*}{\textbf{\#Frame}} & \multicolumn{5}{c|}{\textbf{Spatial Resolution}} & \multirow{2}{*}{\textbf{Max}} \\ 
     ~ & 168 & 196 & 224 & 252 & 280 & ~ \\
    \Xhline{1.0pt}
         4  & \gray{80.4} & \gray{81.7}          & \darkblue{82.3} & \textbf{82.6} & 82.3 & 82.6 \\
     6  & \gray{83.5} & \textbf{84.5}        & 84.3 & 84.2 & 83.6 & 84.5 \\
     8  & 84.4        & \textbf{84.8}        & 84.6 & 84.4 & 83.7 & 84.8 \\
     10 & \textbf{85.2}        & 85.1        & 85.0 & 84.5 & 83.5 & 85.2 \\
     12 & \textbf{85.3}        & 85.3        & 84.9 & 84.4 & 83.4 & 85.3 \\
     16 & \textbf{85.3}        & 85.1        & 84.8 & 84.4 & 83.5 & 85.3 \\
     20 & \textbf{85.1}        & 85.0        & 84.6 & 84.0 & 83.2 & 85.1 \\ \hline
     \textbf{Max} & \textbf{85.3}        & \textbf{85.3}        & \textbf{85.0} & \textbf{84.5} & \textbf{83.7} & - \\
    \Xhline{1.0pt}	
    \end{tabular}
    }
    \vspace{-0.3cm}
    \caption{\textbf{Results of FluxViT-S on K400 using 1024 tokens and different spatiotemporal resolutions.} We use 1clip $\times$ 1crop for testing. The \darkblue{blue} value marks the results of the unmasked setting. The values in bold show the best resolution for each frame count.}
    \label{tab:k400_st_eval_small}
    \vspace{-0.2cm}
\end{table}

\begin{table}[t]
    \centering
    \setlength\tabcolsep{1.1pt}
    \resizebox{0.6\linewidth}{!}{
    \begin{tabular}{l|r|c|c|c}
    \Xhline{1.0pt}	
    \multirow{2}{*}{\textbf{Method}} & \multirow{2}{*}{\textbf{Input Size}}&\multicolumn{3}{c}{\textbf{\#Token}}  \\ 
    ~ & ~ & 1024 & \multicolumn{2}{c}{512} \\	
    \Xhline{0.8pt}
    \multirow{7}{*}{\textbf{Our selector}} & 4$\times$224$^2$ & 82.3 & \multicolumn{2}{c}{79.5} \\ 
     & 8$\times$224$^2$  & 84.6 & \multicolumn{2}{c}{81.3} \\ 
    ~ & 12$\times$224$^2$ & 84.9 & \multicolumn{2}{c}{80.7} \\ 
    ~ & 16$\times$224$^2$ & 84.8 & \multicolumn{2}{c}{80.7} \\
    ~ & 20$\times$224$^2$ & 84.6 & \multicolumn{2}{c}{80.3} \\
    ~ & 24$\times$224$^2$ & 84.6 & \multicolumn{2}{c}{80.3} \\ \cline{2-5}
    ~ & \textbf{Max} & 84.9 & \multicolumn{2}{c}{81.3} \\
    \Xhline{0.8pt}
    \multirow{8}{*}{\textbf{\textit{w/} Vid-TLDR}} & 4$\times$224$^2$ & $\varnothing$ & 77.4 & \gray{78.2} \\ 
    ~ & 8$\times$224$^2$  & 83.9 & 81.0 & \gray{79.8} \\ 
    ~ & 12$\times$224$^2$ & 85.0 & 81.4 & \gray{80.6} \\ 
    ~ & 16$\times$224$^2$ & 85.3 & 81.5 & \gray{80.9} \\
    ~ & 20$\times$224$^2$ & 85.2 & 80.9 & \gray{80.4} \\
    ~ & 24$\times$224$^2$ & 85.2 & 80.5 & \gray{80.3} \\ \cline{2-5}
    ~ & \textbf{Max} & 85.3 & 81.5 & 80.9 \\
    \Xhline{1.0pt}	
    \end{tabular}
    }
    \vspace{-0.3cm}
    \caption{\textbf{Use token merging strategy Vid-TLDR~\cite{vidtldr} on FluxViT K400 testing.} The increment achieved by ViD-TLDR is sensitive to the hyper-parameter setting, like how many tokens are to be reduced in certain layers. }
    \label{tab:vidtldr}
    \vspace{-0.4cm}
\end{table}

\begin{table}[tp]
    \centering
    \small
    \setlength{\tabcolsep}{3pt}
    \resizebox{\linewidth}{!}{
    \begin{tabu}{l|cc|cc|c|c}
        \toprule
        \multirow{2}*{\bf Method} & \multicolumn{2}{c|}{\textbf{K400}} & \multicolumn{2}{c|}{\textbf{K600}} & \multirow{2}*{\textbf{UCF101}} & \multirow{2}*{\textbf{MiTv1}} \\
        ~ & \textbf{Top1} & \textbf{Top5} & \textbf{Top1} & \textbf{Top5} & ~ & ~  \\
        \midrule
        \textbf{FluxViT}-S$_{2048}$ & 66.7 & 88.5 & 65.2 & 87.3 & 85.8 & 28.2 \\ 
        \textbf{FluxViT}-S$_{2048}$+ & 67.0 & 88.8 & 65.5 & 87.5 & 87.5 & 28.6 \\ 
        \textbf{FluxViT}-S$_{1024}$ & 64.2 & 86.6 & 62.8 & 85.7 & 85.1 & 27.2 \\ 
        \textbf{FluxViT}-S$_{1024}$+ & 65.6 & 87.9 & 64.3 & 86.5 & 87.2 & 27.8 \\ 
        \textbf{FluxViT}-S$_{512}$  & 59.0 & 82.6 & 57.4 & 81.2 & 81.5 & 25.5 \\ 
        \textbf{FluxViT}-S$_{512}$+  & 62.6 & 85.5 & 61.1 & 84.2 & 84.2 & 26.5 \\ 
        \midrule
        \textbf{FluxViT}-B$_{2048}$ & 70.2 & 90.6 & 68.9 & 89.5 & 88.7 & 31.2 \\
        \textbf{FluxViT}-B$_{2048}$+ & 70.7 & 90.9 & 69.3 & 89.8 & 89.1 & 31.5 \\
        \textbf{FluxViT}-B$_{1024}$ & 68.6 & 89.6 & 67.2 & 88.4 & 87.8 & 30.4 \\
        \textbf{FluxViT}-B$_{1024}$+ & 69.6 & 90.1 & 68.3 & 89.0 & 89.1 & 31.0 \\
        \textbf{FluxViT}-B$_{512}$  & 64.2 & 86.1 & 62.5 & 84.9 & 84.9 & 28.8 \\
        \textbf{FluxViT}-B$_{512}$+  & 67.4 & 87.4 & 65.8 & 87.7 & 87.6 & 29.8 \\
        \bottomrule
    \end{tabu}
    }
    \vspace{-0.3cm}
    \caption{\textbf{Full Zero-shot Action Recognition Results.}
    }
    \label{tab:more_retrieval_ar}
    \vspace{-0.3cm}
\end{table}
\begin{table}[t!]
    \centering
    \setlength\tabcolsep{6pt}
    \resizebox{\linewidth}{!}{
        \begin{tabular}{l|cc}
        config & SthSth V2 & Others \\
        \Xhline{1.0pt}
        optimizer & \multicolumn{2}{c}{AdamW \cite{adamw}} \\ 
        optimizer momentum & \multicolumn{2}{c}{$\beta_1, \beta_2{=}0.9, 0.98$}  \\
        weight decay & \multicolumn{2}{c}{0.05} \\
        learning rate schedule & \multicolumn{2}{c}{cosine decay~\cite{cosine}} \\
        learning rate & \multicolumn{2}{c}{1e-3}\\
        batch size & \multicolumn{2}{c}{2048} \\
        warmup epochs \cite{warmup} & \multicolumn{2}{c}{20} \\
        total epochs &  \multicolumn{2}{c}{100} \\
        teacher input token & \multicolumn{2}{c}{2048} \\
        student input tokens & \multicolumn{2}{c}{2048, 1536, 1024} \\
        input frame & \multicolumn{2}{c}{(4, 26, stride=2)} \\
        spatial resolution & \multicolumn{2}{c}{(168, 280, stride=28)} \\
        drop path \cite{droppath} & \multicolumn{2}{c}{0.05} \\
        flip augmentation & \textit{no} & \textit{yes} \\
        augmentation & \multicolumn{2}{c}{MultiScaleCrop [0.66, 0.75, 0.875, 1]} \\
        \end{tabular}
    }
    \vspace{-0.3cm}
    \caption{
        \textbf{Flux-UMT pre-training settings.}
    }
    \label{tab:fluxumt_hyperparameters} 
\end{table}
\begin{table}[t!]
    \centering
    \setlength\tabcolsep{2pt}
    \resizebox{0.85\linewidth}{!}{
        \begin{tabular}{l|cc}
        config & Kinetics & COIN \\
        \Xhline{1.0pt}
        optimizer & \multicolumn{2}{c}{AdamW \cite{adamw}} \\ 
        optimizer momentum & \multicolumn{2}{c}{$\beta_1, \beta_2{=}0.9, 0.999$}  \\
        weight decay & \multicolumn{2}{c}{0.05} \\
        learning rate schedule & \multicolumn{2}{c}{cosine decay~\cite{cosine}} \\
        learning rate & \small{2e-4} & \small{5e-4} \\
        batch size & \small{1024+512} & \small{512} \\
        warmup epochs \cite{warmup} & 5+1 & 5 \\
        total epochs & \small{35+5 (S), 20+3 (B)} & \small{40(S), 25 (B)} \\
        drop path \cite{droppath} & \multicolumn{2}{c}{0.1} \\
        flip augmentation & \multicolumn{2}{c}{\textit{yes}} \\
        label smoothing \cite{label_smmoth} & \multicolumn{2}{c}{0.0} \\
        augmentation & \multicolumn{2}{c}{RandAug(9, 0.5) \cite{randaugment}} \\
        \end{tabular}
    }
    \vspace{-0.3cm}
    \caption{
        \textbf{Action recognition fine-tuning settings.} The training epochs A+B on Kinetics include A epochs on K710 and B epochs on K400, the same notation for warmup-epochs and batch size.
    }
    \label{tab:ar_hyperparameters} 
\end{table}
\begin{table}[t!]
    \centering
    \setlength\tabcolsep{6pt}
    \resizebox{0.9\linewidth}{!}{
        \begin{tabular}{l|c}
        config & 25M+2.5M \\
        \Xhline{1.0pt}
        optimizer & AdamW \cite{adamw} \\ 
        optimizer momentum & $\beta_1, \beta_2{=}0.9, 0.98$  \\
        weight decay & 0.02 \\
        learning rate schedule & cosine decay~\cite{cosine} \\
        learning rate & 4e-4 (25M), 2e-5 (2.5M)\\
        batch size & 4096 (image), 4096 (video)\red{$\dag$} \\
        warmup epochs \cite{warmup} & 0.6 (25M), 0 (2.5M)\\
        total epochs &  3 (25M), 1 (2.5M) \\
        input frame & (4, 26, stride=2) \\
        spatial resolution & (168, 280, stride=28) \\
        token threshold & (2048, 4096) \\
        augmentation & MultiScaleCrop [0.5, 1] \\
        \end{tabular}
    }
    \vspace{-0.3cm}
    \caption{
        \textbf{Flux-CLIP pre-training settings.} \red{$\dag$}: For FluxViT-B, we lower the batch size to 2048 for the 2.5M data training.
    }
    \label{tab:stage2_hyperparameters} 
\end{table}
\begin{table}[tp]
    \centering
    \setlength\tabcolsep{1.2pt}
    \resizebox{\linewidth}{!}{
        \begin{tabular}{lccc}
        \toprule
        \textbf{Dataset} & \textbf{\#image/video} & \textbf{\#text} & \textbf{Type} \\
        \midrule
        Kinetics-710 \cite{uniformerv2} & 658K & 0 & Video \\
        COCO \cite{lin2014microsoft} & 113K & 567K & image \\
        Visual Genome \cite{vg} & 100K & 768K & image \\
        SBU Captions \cite{sbu} & 860K & 860K & image \\
        CC3M \cite{cc3m} & 2.88M & 2.88M & image \\
        CC12M \cite{cc12m} & 11.00M & 11.00M & image \\
        S-MiT0.5M \cite{smit} & 0.5M & 0.5M & video \\
        WebVid-2M \cite{bain2021frozen} & 2.49M & 2.49M & video \\
        WebVid-10M \cite{bain2021frozen} & 10.73M & 10.73M & video \\
        InternVid2M \cite{internvid} & 2.0M & 2.0M & video \\
        \midrule
        25M corpus = CC3M$+$CC12M & \multirow{3}{*}{25.68M} & \multirow{3}{*}{26.81M} & \multirow{3}{*}{video + image} \\
         $+$WebVid-10M$+$Visual Genome &~ & ~& ~  \\
         $+$SBU$+$COCO &~ & ~&  ~ \\
        2.5M corpus = S-MiT$+$InternVid2M$+$COCO & 2.56M & 2.62M & video + image \\
        \bottomrule
        \end{tabular}
    }
    \vspace{-0.3cm}
    \caption{\textbf{Statistics of pre-training datasets.}
    }
    \label{tab:statics_pretrain}
    \vspace{-0.1cm}
\end{table}

\begin{table*}
    \centering
    \small
    \setlength{\tabcolsep}{4.0
    pt}
    \resizebox{\textwidth}{!}{
    \begin{tabu}{lr|c|ccc|ccc|ccc|ccc|ccc}
        \Xhline{1.0pt}
        \multirow{2}{*}{\bf Method} & \multirow{2}{*}{\bf \#Token} & \multirow{2}{*}{\bf Type} & \multicolumn{3}{c|}{\bf MSRVTT} & \multicolumn{3}{c|}{\bf DiDeMo} & \multicolumn{3}{c|}{\bf ActivityNet} & \multicolumn{3}{c|}{\bf LSMDC} & \multicolumn{3}{c}{\bf MSVD}\\
        ~ & ~ & ~ &  R@1 & R@5 & R@10 & R@1 & R@5 & R@10 & R@1 & R@5 & R@10 & R@1 & R@5 & R@10 & R@1 & R@5 & R@10\\
        \Xhline{0.8pt}
        \multirow{6}{*}{\ModelName{}-S} & \multirow{2}{*}{2048} & T2V & 44.4 & 67.0 & 75.6 & 48.3 & 74.4 & 82.3 & 52.4 & 79.0 & 87.5 & 20.8 & 36.0 & 44.2 & 49.3 & 77.7 & 85.5 \\
        ~ & ~ & V2T & 44.3 & 67.7 & 77.9 & 50.4 & 75.1 & 83.2 & 53.0 & 79.4 & 88.4 & 21.6 & 37.6 & 45.5 & 78.7 & 92.8 & 95.4 \\
        \cline{3-18}
        ~ & \multirow{2}{*}{1024} & T2V & 42.2 & 64.4 & 74.0 & 45.4 & 72.1 & 79.5 & 47.2 & 73.9 & 84.8 & 18.7 & 35.9 & 43.8 & 47.3 & 76.9 & 84.4 \\
        ~ & ~ & V2T & 43.1 & 65.7 & 74.6 & 47.0 & 71.6 & 80.6 & 48.0 & 75.1 & 85.1 & 20.3 & 37.2 & 45.4 & 79.3 & 92.5 & 95.7 \\
        \cline{3-18}
        ~ & \multirow{2}{*}{512} & T2V & 36.8 & 59.5 & 69.6 & 38.5 & 65.7 & 74.7 & 38.2 & 65.2 & 76.1 & 17.2 & 33.0 & 41.7 & 45.1 & 74.0 & 82.3 \\
        ~ & ~ & V2T & 37.0 & 61.2 & 70.2 & 40.0 & 65.7 & 75.2 & 38.5 & 64.3 & 76.5 & 17.8 & 33.8 & 41.1 & 75.5 & 90.5 & 93.6 \\
        \hline
        \multirow{6}{*}{\ModelName{}-S+} & \multirow{2}{*}{2048} & T2V & 45.0 & 67.5 & 75.8 & 49.2 & 74.5 & 82.8 & 52.4 & 79.0 & 87.5 & 21.1 & 38.2 &   46.0 & 49.7 & 77.8 & 85.8 \\
        ~ & ~ & V2T & 44.9 & 68.2 & 76.5 & 51.2 & 74.9 & 82.9 & 53.8 & 78.0 & 89.2 & 22.4 & 38.6 & 46.4 & 80.2 & 93.6 & 95.5 \\
        \cline{3-18}
        ~ & \multirow{2}{*}{1024} & T2V & 44.5 & 66.4 & 74.6 & 49.0 & 73.9 & 82.4 & 50.3 & 76.9 & 86.4 & 20.5 & 36.6 & 44.8 & 49.1 & 77.0 & 85.5 \\
        ~ & ~ & V2T & 44.2 & 67.4 & 76.4 & 50.5 & 74.1 & 82.4 & 50.9 & 77.8 & 87.3 & 21.7 & 38.5 & 45.3 & 80.2 & 92.2 & 94.5 \\
        \cline{3-18}
        ~ & \multirow{2}{*}{512} & T2V & 40.5 & 62.7 & 71.7 & 45.8 & 71.4 & 80.5 & 44.7 & 71.8 & 82.5 & 19.0 & 34.3 & 40.8 & 46.9 &  76.2 &   84.0  \\
        ~ & ~ & V2T & 41.1 & 63.4 & 73.1 & 47.0 & 71.6 & 79.4 & 44.7 & 71.8 & 82.8 & 19.2 & 35.0 & 41.7  & 78.1 &  90.0 &   93.6 \\
        \hline
        \multirow{6}{*}{\ModelName{}-B} & \multirow{2}{*}{2048} & T2V & 49.8 & 72.2 & 80.1 & 52.2 & 77.5 & 84.5 & 56.6 & 81.5 & 89.6 & 23.7 & 41.0 & 49.3 & 52.6 & 80.1 & 86.7 \\
        ~ & ~ & V2T & 49.3 & 73.6 & 81.5 & 53.0 & 78.9 & 86.7 & 57.6 & 82.9 & 91.3 & 24.8 & 42.0 & 49.3 & 83.3 & 94.2 & 96.6 \\
        \cline{3-18}
        ~ & \multirow{2}{*}{1024} & T2V & 48.0 & 69.6 & 78.0 & 48.8 & 75.5 & 82.6 & 51.8 & 78.4 & 87.5 & 22.6 & 39.9 & 48.3 & 51.9 & 79.5 & 86.2 \\
        ~ & ~ & V2T & 46.5 & 70.5 & 78.0 & 50.5 & 76.4 & 83.5 & 53.4 & 79.7 & 88.4 & 24.0 & 41.4 & 49.1 & 83.0 & 94.8 & 96.9 \\
        \cline{3-18}
        ~ & \multirow{2}{*}{512} & T2V & 42.6 & 64.4 & 73.7 & 42.9 & 68.9 & 77.7 & 42.8 & 69.3 & 79.9 & 20.1 & 36.6 & 45.3 & 49.6  & 77.6 & 84.9 \\
        ~ & ~ & V2T & 41.5 & 65.2 & 74.0 & 44.5 & 70.3 & 77.9 & 43.3 & 70.1 & 80.6 & 21.4 & 37.5 & 46.0 & 80.9 & 92.8 & 95.4 \\
        \hline
        \multirow{6}{*}{\ModelName{}-B+} & \multirow{2}{*}{2048} & T2V & 49.9 & 71.0 & 79.6 & 53.5 &  77.3 & 86.1 & 56.7 & 81.6 & 89.9 & 25.4 & 41.7 & 50.5 & 54.2 & 80.9 & 88.0 \\
        ~ & ~ & V2T & 49.4 & 73.9 & 82.4 & 54.2 & 78.6 & 86.8 & 58.3 & 83.3 & 91.4 & 25.6 & 42.6 & 50.4 & 84.2 & 93.9 & 96.6 \\
        \cline{3-18}
        ~ & \multirow{2}{*}{1024} & T2V & 49.1 & 71.4 & 79.3 & 53.0 & 77.4 & 84.4 & 55.2 & 80.8 & 88.6 & 24.1 & 40.9 & 49.5 & 53.4 & 80.7 & 87.9 \\
        ~ & ~ & V2T & 48.9 & 71.4 & 79.9 & 54.3 & 78.6 & 86.1 & 57.0 & 82.2 & 90.4 & 25.3 & 42.8 & 50.3 & 84.9 & 93.9 & 96.9 \\
        \cline{3-18}
        ~ & \multirow{2}{*}{512} & T2V & 47.2 & 68.6 & 77.0 & 49.8 & 74.6 & 82.4 & 50.3 & 76.0 & 85.0 & 22.5 & 38.2 & 47.1 & 52.1 & 79.3 & 86.7 \\
        ~ & ~ & V2T & 46.5 & 71.1 & 78.6 & 51.2 & 75.5 & 83.4 & 50.9 & 76.7 & 85.6 & 23.0 & 40.3 & 47.6 & 83.0 & 93.9 & 96.4 \\
        \Xhline{1.0pt}
    \end{tabu}
    }
    \vspace{-0.3cm}
    \caption{\textbf{Full Zero-shot retrieval results on MSRVTT, DiDeMo, AcitivityNet, LSMDC, and MSVD.}
    }
    \label{tab:more_retrieval_zs}
\end{table*}
\paragraph{Training dynamics and convergence analysis}

As illustrated in Figure \ref{fig:unstable}, we analyze the gradient norms across the main projector layers in the Flux-Multi Tuned InternVideo2-S. The Patch Embedding Layer exhibits notably elevated gradient norm values, particularly when processing higher input settings with a smaller number of input tokens. This gradient magnitude disparity could potentially introduce training instability. Considering this and to generate more stable token-selection masks as introduced before, we use the Dual Patch Norm module, which shares similar findings with DPN\cite{dpnorm}. However, our convergence analysis, presented in Figure \ref{fig:converg_analysis}, reveals that our Flux-Single tuning with InternVideo2-S, utilizing 3072 tokens and direct tuning over 40 epochs on the K400 dataset, demonstrates consistent convergence patterns. Specifically, configurations with varying frame counts but a fixed token number exhibit normal convergence behavior during tuning. This observation suggests that the aforementioned instability may not be the primary concern in Flux training. Consequently, we prioritize DPN's capability to generate robust masks over its stabilization properties in fine-tuning scenarios and prioritize DPN's capability in stabilized training in the pre-training stage, considering the results in Figure \ref{fig:overall_gnorm} and the results in the previous Flux-UMT ablation study.

\paragraph{Full results using different spatiotemporal resolutions.}

Table \ref{tab:k400_st_eval_small} shows the results of FluxViT-S on K400 using different spatiotemporal resolutions but with a kept 1024 number. Using lower spatial resolution but with a larger frame count can further strengthen the model's performance, which causes another +0.3\% performance gain compared with the best result achieved using standard 224 resolution. This may reflect the dataset's bias towards longer inputs and our method's preference for more dynamic tokens instead of highly informative spatial tokens. Moreover, we find that the best-performing areas are mainly located within a threshold of input tokens, which can be seen as the bolded values of each frame count mainly located within an anti-diagonal line. \textbf{This observation validates our approach of imposing a threshold on input token numbers and suggests an optimized evaluation strategy for determining optimal input configurations.} We propose a systematic evaluation procedure: beginning with minimal input settings (e.g., 4$\times$224$^2$), incrementally increase frame counts until performance plateaus, then progressively reduce spatial resolution while increasing frame count until accuracy improvements cease. This linear complexity evaluation approach efficiently identifies the near-optimal configuration for token optimization. 

\paragraph{Combining modern token merging strategy.}

The integration of state-of-the-art training-free token-merging strategies during inference presents an opportunity to further enhance our Flux method's performance. Table \ref{tab:vidtldr} demonstrates the performance achieved by incorporating the advanced Vid-TLDR \cite{vidtldr} token reduction approach with our FluxViT-S model on K400. Vid-TLDR implements token merging within the initial network blocks to regulate token count. For configurations targeting 1024 tokens, we apply Vid-TLDR to progressively reduce token counts to [2048, 1536, 1024] across the first three layers. Similarly, for 512-token configurations, we evaluate two reduction sequences: [1024, 512] and an alternative [1536, 1024, 512] (denoted by gray values in the table). While results from the first two reduction strategies demonstrate the potential synergy between advanced token-selection methods and our Flux approach, the latter sequence, despite incorporating more tokens in initial layers and involving more computation overhead, underperforms our baseline that employs a heuristic token-reduction method. This outcome shows Vid-TLDR's limitations in accommodating diverse token reduction requirements and highlights its ongoing need for extensive parameter searching. Thus, we only adopt our heuristic but nearly costless token selection method instead of the heavy, nonflexible, and unstable token merging method in Flux.

\subsection{More results}

\paragraph{Full retrieval results.} Table \ref{tab:more_retrieval_zs} shows more zero-shot retrieval results on MSRVTT~\cite{msrvtt}, DiDeMo~\cite{didemo}, ActivityNet~\cite{activitynet}, LSMDC~\cite{lsmdc}, and MSVD~\cite{msvd}. We see that MSRVTT and ActivityNet enjoy only marginal performance gain using 2048 tokens, which may be due to the information saturation for these datasets as also observed in InternVideo2~\cite{iv2} when finding little gain by enlarging the frame count from 4 to 8 and 16. The other three datasets highlight our Flux method's effects more with 2048 tokens, while all these datasets demonstrate our costless performance improvement with 1024 and 512 token inputs.

\paragraph{More zero-shot action recognition results.} Table \ref{tab:more_retrieval_ar} shows more zero-shot retrieval results of our FluxViT on K400~\cite{k400}, K600~\cite{k600}, UCF101~\cite{soomro2012ucf101}, and MiTv1~\cite{mit}. 

\section{More implementation details}

In this section, we introduce the detailed training hyperparameters and report the training dataset details in Table \ref{tab:statics_pretrain}.

\paragraph{Flux-UMT pre-training.} In combining Flux and UMT\cite{umt} framework to get our single modality FluxViT model, we follow most settings as used in deriving InternVideo2 models. Details are shown in Table \ref{tab:fluxumt_hyperparameters}.

\paragraph{Single modality fine-tuning.} We adopt the Flux-UMT pre-trained video encoder and add an extra classification layer for fine-tuning. Input settings are kept the same, and the details of hyperparameters are given in Table \ref{tab:ar_hyperparameters}.

\paragraph{Flux-CLIP per-training.} In combining Flux and CLIP\cite{clip} framework to get our multi-modality FluxViT model, we show the details in Table \ref{tab:stage2_hyperparameters}. We freeze all the modules in 25M data pretraining as Stage 1, except the vision projector. We unfreeze all the modules for the Stage 2 training on the 2.5M dataset. 

\paragraph{Chat Centric Training} We freeze both the LLM and the Vision Encoder in the common stage-1 training of a chat model. We use a learning rate of 1e-3, a batch size of 512, a single training epoch, and a cosine learning rate schedule with a 0.03 warmup rate. 
{
    \small
    \bibliographystyle{ieeenat_fullname}
    \bibliography{main}
}

\end{document}